\documentclass[letterpaper]{article} 
\usepackage{aaai2027}  
\usepackage[hyphens]{url}  
\usepackage{graphicx} 
\usepackage{natbib}  
\usepackage{caption} 
\usepackage{algorithm}
\usepackage{algorithmic}

\usepackage{newfloat}
\usepackage{listings}
\DeclareCaptionStyle{ruled}{labelfont=normalfont,labelsep=colon,strut=off} 
\floatstyle{ruled}
\newfloat{listing}{tb}{lst}{}
\floatname{listing}{Listing}

\usepackage{booktabs}

\usepackage{latexsym}
\usepackage{amssymb}
\usepackage{amsmath}
\usepackage{multirow}
\usepackage[table]{xcolor}
\usepackage{array}
\usepackage{subcaption}
\usepackage{longtable}
\usepackage{makecell}

\newcolumntype{L}[1]{>{\raggedright\arraybackslash}p{#1}}
\newcolumntype{C}[1]{>{\centering\arraybackslash}p{#1}}
\newcolumntype{R}[1]{>{\raggedleft\arraybackslash}p{#1}}

\title{Benchmarking Clinical Decision Pathway Adherence in Large Language Models}
\author{
    Nuo Chen\textsuperscript{\rm 1,6} \quad
    Xinyang Jiang\textsuperscript{\rm 7} \quad 
    Zilong Wang\textsuperscript{\rm 7}\corresponding \quad 
    Zhifei Zhang\textsuperscript{\rm 1} \\ 
    Xiaoye Qu\textsuperscript{\rm 2} \quad 
    Jiajun Deng\textsuperscript{\rm 3} \quad 
    Yulan Guo\textsuperscript{\rm 4,6} \quad 
    Cairong Zhao\textsuperscript{\rm 1,5}\corresponding 
}
\affiliations{
    \textsuperscript{\rm 1}Tongji University \quad
    \textsuperscript{\rm 2}Shanghai Artificial Intelligence Laboratory \quad
    \textsuperscript{\rm 3}Tongji University School of Medicine \\
    \textsuperscript{\rm 4}Sun Yat-sen University \quad
    \textsuperscript{\rm 5}Fuzhou University \quad
    \textsuperscript{\rm 6}Shenzhen Loop Area Institute \quad
    \textsuperscript{\rm 7}Independent Researcher 
}

\begin{document}

\maketitle

\begin{abstract}
Following clinical decision pathways (CDPs) defined by clinical practice guidelines is essential for safe and reliable medical decision-making. However, existing medical large language model (LLM) benchmarks mainly evaluate final-answer accuracy, providing limited evaluation of models' ability to adhere to guidelines. To address this gap, we introduce MEGA-CDP, a benchmark for evaluating whether medical LLMs can generate guideline-adherent CDPs using provided guidelines as references. MEGA-CDP is constructed from 2,274 English and Chinese clinical practice guidelines through a guideline-to-case pipeline, yielding 42,353 clinical cases with explicit reference CDPs. It supports both single-turn vignette and multi-turn interactive settings, and introduces a CDP-oriented evaluation framework for measuring pathway consistency. Experiments on 16 representative LLMs show that reliable clinical decision support remains challenging for current models, demonstrating the need for CDP-oriented evaluation and the value of MEGA-CDP for advancing guideline adherence in medical LLMs.
\end{abstract}


\section{Introduction}
Large language models (LLMs) have shown strong potential in medical applications, motivating a growing body of benchmarks for evaluating their medical knowledge and reasoning capabilities \cite{medqa,medmcqa,pubmedqa}. Meanwhile, LLMs are increasingly being explored for clinical decision-making tasks, including diagnosis, treatment planning, and clinical decision support \cite{cdaflow, vatsal2026agentic}. Unlike isolated question answering, clinical decision-making requires multi-step reasoning over heterogeneous patient evidence. This difference highlights the need to evaluate not only final decisions, but also the processes by which they are produced.

\begin{figure*}[htbp]
    \centering
    \includegraphics[width=\textwidth]{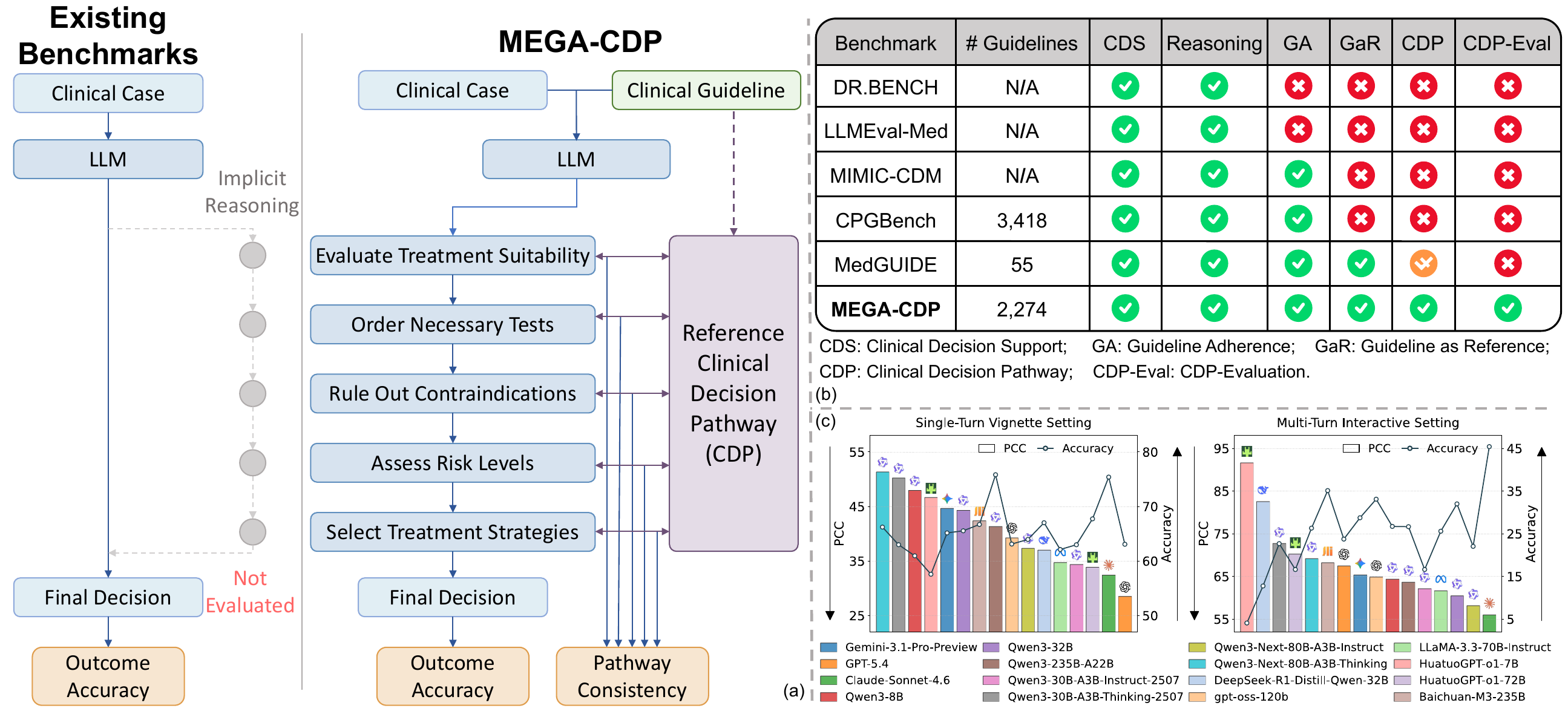}
    \caption{Overview of MEGA-CDP. 
(a) Comparison between existing benchmarks and MEGA-CDP. 
(b) Feature comparison between MEGA-CDP and representative clinical decision-making benchmarks. 
(c) Representative model performance under the two settings.}
    \label{fig:teaser}
\end{figure*}


In clinical practice, safe and reliable decision-making requires strict adherence to clinical decision pathways (CDPs) defined by established guidelines \cite{tso2017automating}. As illustrated in Figure~\ref{fig:teaser}(a), a CDP consists of ordered intermediate steps whose omission or misordering can lead to decisions that appear correct but are clinically unsafe, non-explainable, or difficult to review \cite{ng2025clinical, mienye2024survey}. This highlights the necessity of CDP adherence for medical LLMs. Moreover, guideline-defined CDPs may vary across regions, healthcare resources, and clinical contexts, and they are frequently updated as medical evidence evolves \cite{steinberg2011clinical}, requiring models to reason with provided guidelines as references rather than relying solely on internal parametric knowledge. However, existing works largely overlook these two aspects, focusing primarily on final outcomes or general rationale quality. This motivates our goal of evaluating whether LLMs can use provided guidelines as references to generate clinical decisions that faithfully follow guideline-defined CDPs.

To address this limitation, we introduce the \textbf{Me}dical \textbf{G}uideline-\textbf{A}dherent \textbf{C}linical \textbf{D}ecision \textbf{P}athway (MEGA-CDP), a large-scale benchmark for systematically evaluating the ability of LLMs to generate clinical decisions that adhere to guidelines. Specifically, MEGA-CDP comprises 2,274 real-world clinical practice guidelines collected from publicly available English and Chinese sources between 2020 and 2025. We develop a rigorous and automated pipeline that transforms the textual content of each guideline into a reasoning tree, from which guideline-defined CDPs are extracted. The pipeline then synthesizes corresponding clinical cases by traversing each CDP. In this design, each case can be correctly resolved using the full guideline text, which provides all necessary information for decision-making without requiring external knowledge. This formulation yields structured, pathway-aligned data that supports fine-grained evaluation of guideline adherence, while remaining scalable for continuous expansion and updates as real-world guidelines evolve. The resulting dataset contains over 42,353 curated CDPs and corresponding clinical cases.

Evaluating predicted CDPs is challenging because they may differ in length from the reference CDPs, intermediate steps can be expressed in diverse natural language forms, and models may omit necessary steps, introduce redundant steps, or reorder steps, potentially violating the intended decision logic. Currently, there is no standardized method for assessing whether model-generated CDPs align with the ground-truth reference. To tackle this challenge, we propose a CDP-oriented evaluation framework that first estimates step-level consistency between predicted and reference steps and then aggregates them to compute pathway-level consistency using dynamic time warping. As shown in Figure~\ref{fig:teaser}(c), we applied this evaluation framework to 16 representative LLMs under two task settings: (i) a single-turn vignette setting, where the model produces a full CDP from a summarized clinical case, and (ii) a multi-turn interactive setting, where the model incrementally acquires information and updates its decisions. This evaluation provides fine-grained insights into the models' ability to produce guideline-adherent CDPs.

\begin{figure*}[htbp]
    \centering
    \includegraphics[width=\textwidth]{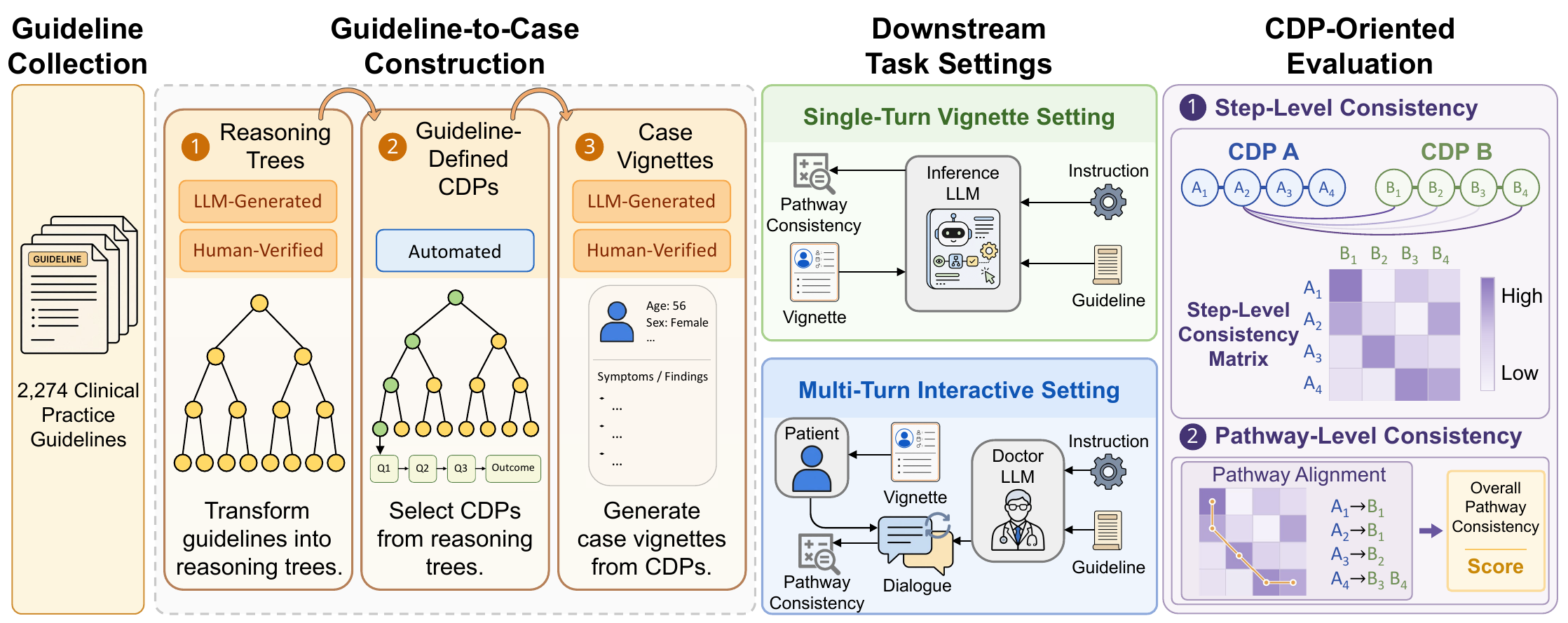}
    \caption{Framework of MEGA-CDP. The framework consists of guideline collection, guideline-to-case construction, downstream task settings, and CDP-oriented evaluation. Clinical guidelines are transformed into reasoning trees, guideline-defined CDPs, and case vignettes to support both single-turn vignette and multi-turn interactive settings. Model-generated CDPs are then evaluated through a two-stage procedure for measuring pathway consistency.}
    \label{fig:overall}
\end{figure*}

\section{Related Work}
Prior work has introduced a range of benchmark datasets for evaluating medical question answering systems. Representative examples include MedQA \cite{medqa}, MedMCQA \cite{medmcqa}, and PubMedQA \cite{pubmedqa}, which are formulated as exam-style tasks. Other benchmarks extend this setting to more context-rich scenarios. For instance, emrQA \cite{emrqa} introduces question answering over electronic medical records, while MedDialog \cite{meddialog} incorporates large-scale doctor–patient conversations. CliMedBench \cite{climedbench} further expands the scope by covering multiple clinical scenarios, tasks, and evaluation metrics. These benchmarks evaluate medical domain knowledge through multiple-choice or open-ended question formats.

Recent efforts have explored the evaluation of clinical decision-making and reasoning processes through dedicated benchmarks. DR.BENCH \cite{drbench} introduces diagnostic reasoning tasks for assessing clinical decision-making. MedJourney \cite{medjourney} evaluates decision-making in multi-stage scenarios with progressively revealed patient information, capturing the dynamic nature of clinical decision processes. LLMEval-Med \cite{llmevalmed} adopts an LLM-as-a-Judge \cite{zheng2023judging} paradigm to enable automated evaluation of reasoning quality. Building on this trend toward reasoning-focused evaluation, subsequent studies have emphasized guideline-based clinical assessment. MIMIC-CDM \cite{mimiccdm} evaluates an LLM's ability to generate decisions consistent with authoritative clinical guidelines, while CPGBench \cite{cpgbench} examines whether LLMs can provide guideline-adherent recommendations in multi-turn conversations. More closely related to our work, MedGUIDE \cite{medguide} measures adherence to CDPs explicitly defined by existing guideline flowcharts.

Current studies have begun to examine multi-step reasoning and sequential decision-making in clinical settings, exploring methods that simulate diagnostic workflows, treatment planning, or agent-driven reasoning \cite{cdaflow, medagents, clinicalagent, medagentsbench}. These investigations underscore the critical role of stepwise information integration and decision alignment in realistic medical scenarios. Yet, most existing evaluation frameworks do not explicitly encode guideline-based decision structures, nor do they systematically assess whether generated reasoning follows CDPs. As shown in Figure~\ref{fig:teaser}(b), addressing this limitation, MEGA-CDP provides a CDP-oriented benchmark that represents guidelines as CDPs, linking them to structured case vignettes and canonical reasoning sequences, which enables rigorous process assessment of guideline adherence and the fidelity of model-generated reasoning.

\section{MEGA-CDP}
\subsection{Data Curation Pipeline}
We design a two-stage data curation pipeline that first extracts guideline-defined CDPs from clinical practice guidelines and then synthesizes CDP-aligned case vignettes for downstream evaluation.

\noindent
\textbf{Guideline Collection.}
We collected clinical practice guidelines published between 2020 and 2025 from publicly available English and Chinese sources. Text-based guidelines were prioritized, while those dominated by explicit flowcharts or diagram-based representations were excluded, resulting in 2,274 retained guidelines. To facilitate downstream construction, we converted the guideline PDFs into structured Markdown format using PaddleOCR \cite{cui2026paddleocrvl15multitask09bvlm} for text extraction and Ovis2.5-9B \cite{lu2025ovis25technicalreport} for layout understanding and formatting.

\noindent
\textbf{Guideline-to-Case Construction.}
\label{sec:DataConstruction}
As shown in Figure~\ref{fig:overall}, we construct the MEGA-CDP dataset through a multi-stage automated pipeline implemented with GPT-5.2 \cite{openai2026gpt52}. The construction process consists of two steps:

\textbf{(1) Clinical Decision Pathway (CDP) Extraction.}
Given the raw guideline text, we use LLMs with structured prompting templates to extract guideline-defined CDPs. The extracted decision logic is represented as a tree structure, where each non-leaf node corresponds to a clinical decision condition, each branch represents a possible outcome of the corresponding condition and leads to subsequent decision steps, and each leaf node represents a final decision outcome. 
We then enumerate all root-to-leaf paths, yielding a total of 42,353 guideline-defined CDPs. Each CDP contains an ordered sequence of decision conditions leading to a specific outcome, and therefore serves as the reference CDP for subsequent evaluation. These CDPs represent guideline-defined reasoning pathways rather than an exhaustive set of all possible clinical workflows, and are designed to evaluate whether models can reproduce the decision logic specified by guidelines. By converting free-text guideline recommendations into explicit CDPs, the pipeline provides a structured basis for constructing cases whose correct decisions can be traced back to guideline-defined reasoning steps.

\textbf{(2) Case Vignette Synthesis.}
Given each extracted CDP, we synthesize a corresponding clinical case vignette with LLMs. The prompt instructs the model to transform the CDP into a case that is consistent with the pathway and leads to a uniquely determined terminal outcome.
The resulting case vignette integrates the information required by the CDP into a natural clinical narrative without explicitly exposing the underlying decision structure. In this way, each vignette is aligned with one reference CDP and contains the necessary information for reaching its associated decision outcome. 

\noindent
\textbf{Data Quality Control.}
To ensure the reliability of the constructed data, we conducted quality control through LLM-based sampling inspection and human review. Specifically, generated instances were sampled and examined for consistency among the reasoning tree, CDP, and case vignette. Cases with obvious contradictions, unsupported decision logic, or inconsistencies between the case description and the corresponding pathway were excluded from the dataset. 
Across the CDPs and case vignettes derived from 500 sampled guidelines, fewer than 0.5\% of the CDPs and fewer than 1\% of the case vignettes were identified as problematic.

\subsection{Downstream Task Settings}
\label{sec:TaskSetting}
To evaluate the decision-making processes of LLMs, as shown in Figure~\ref{fig:overall}, we propose two downstream task settings.

\noindent
\textbf{Single-Turn Vignette Setting.} 
In the single-turn vignette setting, the input consists of a clinical guideline and a corresponding case vignette. The model is then instructed to follow the guideline and produce a step-by-step predicted CDP together with the final decision outcome. To ensure a consistent descriptive style for CDP comparison, we convert the reference CDP into a sequence of concise reasoning narratives in this setting. This setting evaluates whether the model can derive a complete guideline-adherent CDP from summarized clinical information.

\noindent
\textbf{Multi-Turn Interactive Setting.} 
In the multi-turn interactive setting, the model maintains access to the clinical guideline and dialogue history while interacting with a patient-side environment. At each turn, the model asks one question to acquire additional patient information, after which the environment responds according to the clinical case. The interaction continues until the model produces a final decision outcome. Afterwards, the sequence of generated questions is treated as the predicted CDP for evaluation. This setting evaluates whether the model can actively acquire patient information in a way that follows the guideline-defined CDP.

\subsection{CDP-Oriented Evaluation}
We propose a CDP-oriented evaluation method that compares the predicted CDPs generated in downstream task settings with the reference CDPs constructed during data construction. As illustrated in Figure~\ref{fig:overall}, we first assess step-level consistency between predicted and reference CDP steps, and subsequently aggregate the resulting scores to measure pathway-level consistency.

\noindent
\textbf{Step-Level Consistency Scoring.}
\label{sec:step_level_consistency}
We evaluate consistency at the step level using an LLM-as-a-judge paradigm. Given the task instruction, the predicted CDP, and the reference CDP, the judge model compares each predicted step with the reference steps by assessing whether it covers the same key decision point and preserves the corresponding guideline-defined logic. Following G-Eval \cite{liu2023g}, the judge model assigns a score on a 5-point Likert scale, and the scores are averaged over five repeated runs to improve reliability. We then normalize the averaged scores to the range of $[0,1]$ for subsequent pathway-level measurement. Before formal evaluation, we iteratively refined the scoring prompt to improve its agreement with human review. The resulting step-level scores capture fine-grained alignment between the predicted and reference CDPs and serve as the basis for pathway-level consistency measurement.

We considered three candidate judge models, including GPT-5.4 Mini \cite{openai2026gpt54mini}, Qwen3-8B \cite{qwen3technicalreport}, and BERTScore \cite{zhang2019bertscore}, and conducted experiments to compare their scoring behavior. Based on the threshold analysis in Section~\ref{sec:threshold_analysis}, GPT-5.4 Mini showed better alignment with human judgments on the sampled instances, and was therefore selected as the judge model for the main evaluation. Subsequently, we conducted a correlation analysis to assess the agreement among the candidate judge models.


\begin{table*}[t]
\centering
\small
\begin{tabular}{L{5cm}|C{1.4cm}C{1.4cm}|C{1.4cm}C{1.4cm}C{1.4cm}}
\toprule
\multirow{2}{*}{Model} 
& \multicolumn{2}{c|}{Single-Turn} 
& \multicolumn{3}{c}{Multi-Turn} \\
& PCC (\%) $\downarrow$
& Acc (\%) $\uparrow$
& PCC (\%) $\downarrow$
& Acc (\%) $\uparrow$
& SR (\%) $\uparrow$\\
\midrule
\rowcolor{gray!15}
\multicolumn{6}{c}{\textit{Proprietary Models}} \\
Gemini-3.1-Pro-Preview          & 44.61          & 65.14         & 65.32         & 28.73         & 78.17\\
GPT-5.4                         & \textbf{28.41} & 63.06         & 67.43         & 23.73         & 74.77\\
Claude-Sonnet-4.6               & 32.42          & \textbf{75.38}& \textbf{55.92}& \textbf{45.40}& \textbf{99.22}\\

\midrule
\rowcolor{gray!15}
\multicolumn{6}{c}{\textit{Open-source Models}} \\
Qwen3-8B                        & 47.88         & 60.94          & 64.34            & 26.70          & 97.65\\
Qwen3-32B                       & 44.31         & 65.53          & 60.45            & 31.98          & 99.24\\
Qwen3-235B-A22B                 & 41.23         & \textbf{75.81} & 63.66            & 26.64          & 95.44\\
Qwen3-30B-A3B-Instruct-2507     & \textbf{34.29}& 62.97          & 61.95            & 16.61          & 97.44\\
Qwen3-30B-A3B-Thinking-2507     & 50.17         & 62.98          & 72.61            & 22.67          & 76.21\\
Qwen3-Next-80B-A3B-Instruct     & 37.26         & 63.94          & \textbf{58.27}   & 22.03          & 98.71\\
Qwen3-Next-80B-A3B-Thinking     & 51.31         & 66.21          & 69.15            & 26.30          & 80.01\\
DeepSeek-R1-Distill-Qwen-32B    & 36.98         & 67.01          & 82.12            & 12.75          & 58.69\\
gpt-oss-120b                    & 39.24         & 63.10          & 64.78            & \textbf{33.08} & 91.85\\
LLaMA-3.3-70B-Instruct          & 34.71         & 62.10          & 61.04            & 25.58          & \textbf{99.77}\\

\midrule
\rowcolor{gray!15}
\multicolumn{6}{c}{\textit{Specialized Medical Models}} \\
HuatuoGPT-o1-7B                 & 46.56         & 57.56         & 91.73         & 4.07          & 24.05\\
HuatuoGPT-o1-72B                & \textbf{33.87}& \textbf{67.72}& 70.28         & 16.61         & 74.68\\
Baichuan-M3-235B                & 42.35         & 66.69         & \textbf{68.27}& \textbf{35.10}& \textbf{98.87}\\

\bottomrule
\end{tabular}  
\caption{\label{tab:results}
Performance of evaluated LLMs on single-turn vignette and multi-turn interactive settings. PCC denotes the pathway consistency cost, Acc denotes the outcome accuracy, and SR denotes the success rate. Arrows indicate whether higher ($\uparrow$) or lower ($\downarrow$) values are better.
  }
\end{table*}

\noindent
\textbf{Pathway-Level Consistency Measurement.}
Building on the step-level consistency scores, we compute a Pathway Consistency Cost (PCC) between the predicted CDP and the reference CDP to assess the extent to which the model's decision-making process adheres to the guideline. We design three candidate methods for pathway-level consistency measurement, including Dynamic Time Warping (DTW), Optimal Transport (OT), and Longest Non-Decreasing Subsequence (LNDS). Based on empirical performance, we select DTW as the primary metric for the main evaluation. The algorithmic details of the alternative metrics, OT and LNDS, are provided in the supplementary materials.

In our benchmark setting, we evaluate whether models can follow guideline-defined CDPs. A model may omit intermediate decisions, introduce additional steps, or deviate from the decision order specified by the guideline, resulting in misalignments between the predicted and reference pathways. Therefore, we formulate pathway comparison as a sequence alignment problem and employ the DTW method, which identifies the optimal alignment between two ordered sequences of potentially different lengths while preserving their sequential order. Let $m$ and $n$ denote the number of steps in the reference and predicted CDPs, respectively. After step-level consistency scoring, the scores form a consistency matrix $C\in\mathbb{R}^{m\times n}$, where each entry $C_{i,j}$ denotes the consistency score between the $i$-th reference step and the $j$-th predicted step. We then convert the consistency matrix into a distance matrix $D\in\mathbb{R}^{m\times n}$, which is used as the cost matrix in the DTW formulation, by defining $D_{i,j}=1-C_{i,j}$ so that larger values correspond to higher alignment costs.

Using this cost matrix, the alignment is represented by a path $\mathcal{P}=\{(i_k,j_k)\}_{k=1}^{K}$, where each pair $(i_k,j_k)$ indicates that the $i_k$-th reference step is aligned with the $j_k$-th predicted step. Here, $K$ denotes the total length of the alignment path. The alignment path satisfies the standard DTW constraints, including boundary conditions, monotonicity, and continuity, so that the relative order of decision steps is preserved. The DTW objective is to find the alignment path $\mathcal{P}$ that minimizes the accumulated alignment cost:
\begin{equation}
    \mathcal{P}^* = \operatorname*{argmin}_{\mathcal{P}} \sum_{k=1}^{K} D_{i_k,j_k}.
\end{equation}

The DTW algorithm enables flexible alignment under sequence variation, allowing one-to-many or many-to-one correspondences between steps while preserving the global ordering of the CDP. As the final metric, we normalize the accumulated alignment cost by the length of the alignment path as
\begin{equation}
    \mathrm{PCC} = \frac{1}{K}\sum_{k=1}^{K}D_{i_k,j_k}.
\end{equation}
The PCC reflects the loss of order alignment between the reference CDP and the predicted CDP, with lower values indicating better alignment.

We analyze the agreement between the primary DTW-based metric and the two alternative metrics in Section~\ref{sec:correlation}. Additionally, a human preference study is conducted to evaluate whether PCC aligns with human judgments, with details provided in the supplementary materials.

\section{Experiments}

\subsection{Experimental Setup}
To evaluate LLMs' ability to perform guideline-adherent clinical decision-making, we selected three types of models: (1) proprietary models, including the GPT series \cite{openai2026gpt54}, Claude series \cite{anthropic2026claudesonnet46systemcard}, and Gemini series \cite{google2026gemini31propreview}; (2) open-source models, including the DeepSeek series \cite{deepseekai2025deepseekr1incentivizingreasoningcapability} , Qwen series \cite{qwen3technicalreport}, gpt-oss series \cite{openai2025gptoss120bgptoss20bmodel}, and LLaMA series \cite{grattafiori2024llama}; and (3) specialized medical models, including the HuatuoGPT series \cite{chen2024huatuogpto1medicalcomplexreasoning} and Baichuan series \cite{dou2026baichuan}. Given that MEGA-CDP involves long clinical guidelines in multiple languages, we restricted our evaluation to models that support both English and Chinese inputs and provide a context window of at least 40K tokens. All models were evaluated with their default configurations.

We constructed the test set from 440 clinical practice guidelines sampled from the collected corpus, including 210 Chinese guidelines and 230 English guidelines, yielding 7,458 cases for evaluation. Each test instance consisted of a clinical guideline, a case vignette, the corresponding final decision outcome, and reference CDPs for evaluation. We used the same underlying test instances for both the single-turn vignette setting and the multi-turn interactive setting, while adapting the input format and task instruction to each setting. In addition to PCC, we report outcome accuracy for all models to measure the correctness of the final clinical decision. For the multi-turn setting, we also report the success rate, defined as the proportion of cases in which the model follows the required interaction format and successfully completes the multi-turn dialogue.

\subsection{Main Results}
Table~\ref{tab:results} summarizes the performance of all evaluated models under the single-turn vignette and multi-turn interactive settings. Overall, model performance varies substantially across model families and task settings. Proprietary models generally achieve the strongest overall performance, while general open-source models provide acceptable alternatives. In contrast, specialized medical models do not consistently show advantages in understanding and following medical guidelines. Notably, although Baichuan-M3-235B is fine-tuned from Qwen3-235B-A22B and has been further trained to model clinical decision-making processes, its PCC performance degrades under both settings. This suggests that medical specialization does not necessarily translate into better guideline-adherent reasoning and may even compromise the model's underlying reasoning ability. We also observe that, under the same model family and parameter scale, thinking variants may exhibit weaker pathway consistency than their instruct counterparts, as seen in Qwen3-30B-A3B and Qwen3-Next-80B-A3B. These results provide new insights into the generation of guideline-adherent CDPs in clinical decision-making.
Additional results for the candidate metrics and judge models are reported in supplementary materials.

\begin{figure*}[t]
    \centering
    \includegraphics[width=\linewidth]{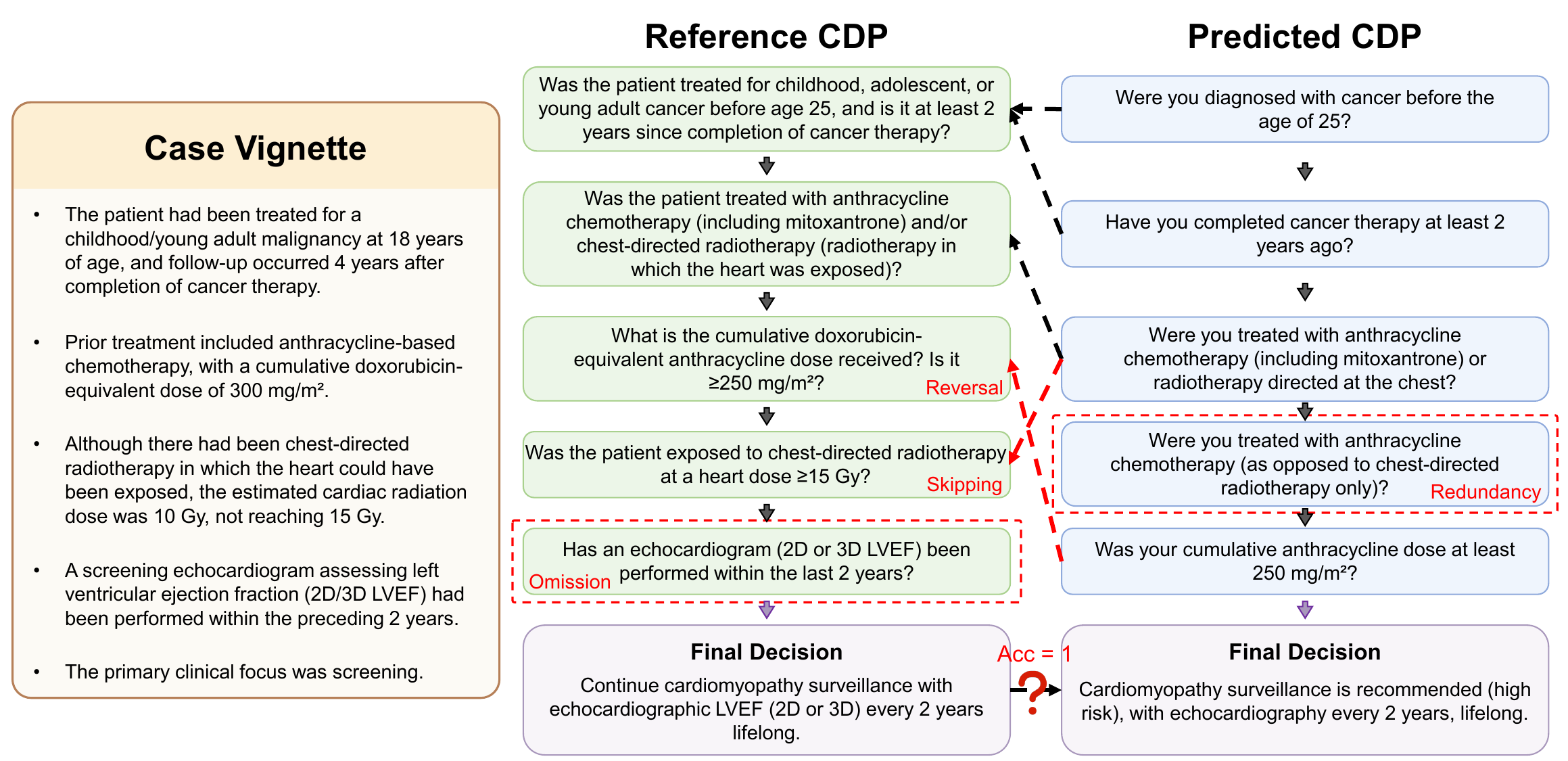}
    \caption{Example of pathway-level inconsistency in guideline-adherent reasoning. Although the model reaches the correct final decision, its predicted CDP still deviates from the reference CDP through reversal, skipping, omission, and redundancy.}
    \label{fig:qualitative}
\end{figure*}

\noindent
\textbf{Analysis of the Single-Turn Vignette Setting.}
In the single-turn vignette setting, GPT-5.4 achieves the best pathway consistency with the lowest PCC of 28.41\%. Qwen3-235B-A22B obtains the highest outcome accuracy of 75.81\%, while Claude-Sonnet-4.6 offers the best overall trade-off, with a PCC of 32.42\% and an accuracy of 75.38\%. More broadly, pathway consistency and outcome accuracy do not always change consistently across models. Some models achieve high accuracy despite relatively higher PCC, showing that correct final decisions do not necessarily imply correct reasoning processes. This highlights the necessity of process-level evaluation beyond outcome accuracy.

We further observe that model scale, model version, and medical fine-tuning all affect single-turn performance. Larger models within the same family generally perform better, but improvements across generations are not strictly monotonic. Medical fine-tuning can help in some cases, but its effect is not consistent across medical models. These findings suggest that guideline-adherent reasoning depends on multiple factors beyond final-answer capability.

\noindent
\textbf{Analysis of the Multi-Turn Interactive Setting.}
The multi-turn interactive setting is substantially more challenging than the single-turn vignette setting. Success rates vary considerably across models, reflecting differences in their ability to follow interaction instructions. Compared with the single-turn setting, PCC increases by 26.98 percentage points on average, while outcome accuracy decreases by 40.51 percentage points. This overall performance degradation is consistent with prior findings on the difficulty of multi-turn clinical decision-making~\cite{mimiccdm,laban2025llms}. Claude-Sonnet-4.6 achieves the best overall performance in this setting, with the lowest PCC of 55.92\% and the highest outcome accuracy of 45.40\%.

Model-generated CDPs show more severe process deviations in the multi-turn setting. Figure~\ref{fig:qualitative} shows an example from Claude-Sonnet-4.6, the best-performing model in the multi-turn setting. Although the model reaches the correct final decision, its predicted CDP still deviates from the reference CDP. In this example, the predicted CDP exhibits four common types of process deviations, including omission, redundancy, skipping, and reversal. These errors indicate that multi-turn clinical inquiry remains challenging, as current LLMs may reach the correct outcome while still failing to produce a guideline-adherent CDP.

Specialized medical models also show inconsistent effects in this setting. Baichuan-M3-235B achieves better outcome accuracy and success rate than similar-scale general models, but its PCC is worse. This may reflect a mismatch between its clinical decision-making fine-tuning and our evaluation setup, where models must generate CDPs by using the provided guideline as a reference. In contrast, HuatuoGPT-o1 degrades more broadly across multi-turn metrics. These results suggest that current medical fine-tuning does not consistently improve the generation of guideline-adherent CDPs.

\subsection{Agreement between PCC and Human Preferences}
\label{sec:pcc_human}
To assess whether PCC aligns with human judgments, we conduct a pairwise human preference study. Given a reference pathway and two model-generated pathways, two human annotators independently select the pathway more consistent with the reference. We compare aggregated human preferences with PCC-based pairwise decisions and calculate agreement rates.

We randomly sample 100 pathway pairs from the multi-turn setting results across different models. Since low-quality model outputs may introduce ambiguity into human judgments, we additionally sample another 100 pairs from two relatively strong-performing models, Claude-Sonnet-4.6 and Qwen3-Next-80B-A3B-Instruct. Both groups contain equal numbers of English and Chinese cases.

Our human preference study shows strong agreement between PCC and human judgments, with agreement rates of 88.5\% and 91.5\% for the two groups, respectively, and an overall agreement rate of 90.0\% across all 200 pairs. The agreement between the two human annotators is also 90.0\%, indicating that PCC achieves a level of consistency comparable to human agreement. These results suggest that PCC provides a reliable measure of pathway consistency that is aligned with human judgments.

\subsection{Threshold Analysis for Step-Level Consistency Scores}
\label{sec:threshold_analysis}
Step-level consistency scores provide a graded measure of how well a predicted step matches a reference step. However, different judge models may exhibit different calibration behaviors when assessing consistency, making their scores not directly comparable under a single universal threshold, even though the scoring criteria are explicitly defined in the prompt. We therefore design an analysis to determine appropriate thresholds for step-level matching that align judge model scores with human judgments.

\begin{figure}[t]
    \centering
    \includegraphics[width=\linewidth]{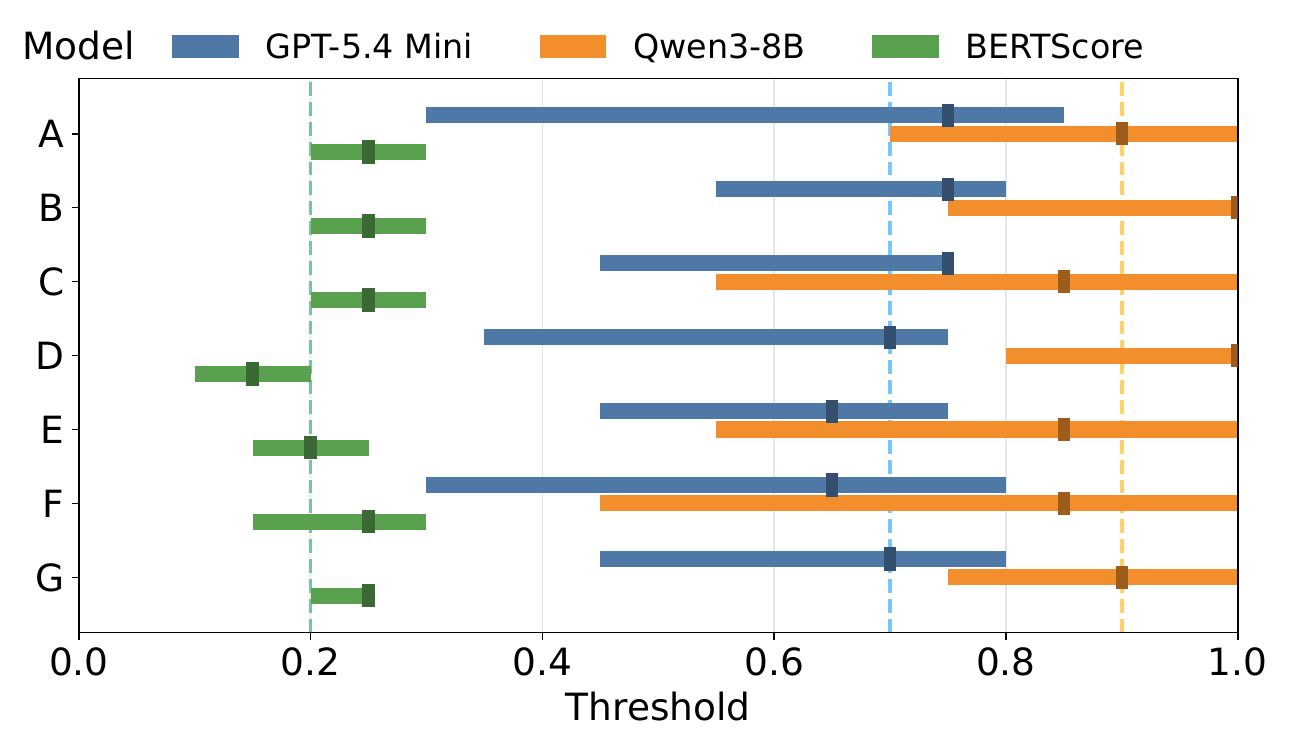}
    \caption{Threshold experiment results for step-level matching. The seven models labeled A through G in the figure are, respectively, Qwen3-Next-80B-A3B-Instruct, Qwen3-Next-80B-A3B-Thinking, Qwen3-8B, Qwen3-235B-A22B, HuatuoGPT-o1-7B, HuatuoGPT-o1-72B, and Baichuan-M3-235B.}
    \label{fig:threshold}
\end{figure}

Specifically, we sample prediction results from seven representative models and perform human annotation. For each reference step in the model predictions, we obtain human judgments on whether its best-matched predicted step corresponds to a correct match. We then compute the F1 score of the binary matching results under a range of candidate thresholds at fixed intervals. Around the threshold that achieves the highest F1 score, we identify a stable region as the candidate threshold range. The results are shown in Figure~\ref{fig:threshold}. Considering both the optimal threshold and the common stable region, we set the matching thresholds to 0.7 for GPT-5.4 Mini, 0.9 for Qwen3-8B, and 0.2 for BERTScore.

\begin{table}[t]
\centering
\small
\begin{tabular}{c|c|ccc|c}
\toprule
Setting & Pair & $r$ & $\rho$ & $\tau$ &$W$ \\
\midrule
\multirow{3}{*}{DTW} 
    & GPT vs Qwen & 0.79 & 0.79 & 0.60 & \multirow{3}{*}{0.70} \\
    & BERT vs GPT & 0.48 & 0.46 & 0.32 &  \\
    & Qwen vs BERT & 0.39 & 0.38 & 0.26 &  \\
\midrule
\multirow{3}{*}{OT} 
    & GPT vs Qwen & 0.80 & 0.79 & 0.60 & \multirow{3}{*}{0.71} \\
    & BERT vs GPT & 0.50 & 0.49 & 0.34 &  \\
    & Qwen vs BERT & 0.43 & 0.43 & 0.29 &  \\
\midrule
\multirow{3}{*}{LNDS} 
    & GPT vs Qwen & 0.72 & 0.72 & 0.58 & \multirow{3}{*}{0.71} \\
    & BERT vs GPT & 0.52 & 0.53 & 0.41 &  \\
    & Qwen vs BERT & 0.47 & 0.47 & 0.36 &  \\
\midrule
\midrule
\multirow{3}{*}{GPT} 
    & DTW vs OT & 0.88 & 0.86 & 0.70 & \multirow{3}{*}{0.87} \\
    & LNDS vs DTW & 0.82 & 0.82 & 0.66 &  \\
    & OT vs LNDS & 0.76 & 0.75 & 0.59 &  \\
\midrule
\multirow{3}{*}{Qwen} 
    & DTW vs OT & 0.86 & 0.83 & 0.66 & \multirow{3}{*}{0.84} \\
    & LNDS vs DTW & 0.73 & 0.73 & 0.57 &  \\
    & OT vs LNDS & 0.71 & 0.71 & 0.55 &  \\
\midrule
\multirow{3}{*}{BERT} 
    & DTW vs OT & 0.94 & 0.93 & 0.79 & \multirow{3}{*}{0.84} \\
    & LNDS vs DTW & 0.71 & 0.71 & 0.54 &  \\
    & OT vs LNDS & 0.69 & 0.68 & 0.51 &  \\

\bottomrule
\end{tabular}  
\caption{\label{tab:correlation}
Correlation analysis of metrics and judge models. Pearson $r$, Spearman $\rho$, Kendall $\tau$, and Kendall's $W$ are reported for pairwise comparisons. The table is divided into groups fixing either the metric or the judge model.}
\end{table}

\subsection{Correlation Analysis of Metrics and Judge Models}
\label{sec:correlation}
To validate the generality of our evaluation framework, we perform a correlation analysis across different pathway-level consistency metrics and judge models. Since LNDS has opposing directionality relative to DTW and OT, its scores are inverted. Table~\ref{tab:correlation} reports Pearson correlation $r$, Spearman rank correlation $\rho$, and Kendall rank correlation $\tau$ for pairwise comparisons, as well as overall agreement measured by Kendall's $W$.

When fixing the metric, GPT-5.4 Mini and Qwen3-8B, as candidate LLM-as-a-judge models, exhibit strong correlations across all metrics, whereas BERTScore shows moderate agreement with the other two. When fixing the judge model, DTW and OT, which leverage global alignment information, show higher pairwise consistency, while LNDS also demonstrates reasonable agreement. Both groups achieve high Kendall's $W$, suggesting that the evaluation results remain largely consistent across metric and judge model choices.

These results demonstrate the feasibility and robustness of the selected evaluation components. Based on this analysis, we choose GPT-5.4 Mini as the default judge model and DTW as the primary metric, while also providing alternative judge models and metrics for experimental use.

\section{Conclusion}
In this paper, we introduced MEGA-CDP, a large-scale benchmark for evaluating whether medical LLMs can generate guideline-adherent CDPs. MEGA-CDP includes an automated guideline-to-case construction pipeline that extracts guideline-defined CDPs from real-world guidelines and synthesizes pathway-aligned clinical cases. Built on this pipeline, MEGA-CDP supports both single-turn vignette and multi-turn interactive settings with a CDP-oriented evaluation framework. Experiments on 16 representative LLMs show that outcome accuracy and pathway consistency capture different aspects of clinical decision-making, that both task settings remain substantially challenging, and that specialized medical models do not consistently outperform general-purpose models. These findings highlight the importance of CDP-oriented evaluation and demonstrate the value of MEGA-CDP for developing more reliable and guideline-adherent medical LLMs.

\section*{Acknowledgments}
This work was supported by the Shanghai Municipal Special Program for Basic Research on General AI Foundation Models (Grant No. 2025SHZDZX025G10), in collaboration with Shanghai Artificial Intelligence Laboratory.
This work was supported by the Shenzhen Loop Area Institute under grant FPF10120260001

\bibliography{aaai2027}


\clearpage
\appendix

\begin{table}[t]
\small
\centering
\begin{tabular}{l|l}
\toprule
Component & Specification \\
\midrule
\rowcolor{gray!15}
\multicolumn{2}{c}{\textit{Hardware Infrastructure}} \\
GPU & 8$\times$ NVIDIA H100 80GB \\
CPU & Intel Xeon Platinum 8468V \\
Memory & 1920 GB \\
\midrule
\rowcolor{gray!15}
\multicolumn{2}{c}{\textit{Software Environmen}} \\
Operating System & Ubuntu 20.04 LTS \\
CUDA & 12.6 \\
Python & Python 3.10 \\
PyTorch & 2.9.1 \\
SGLang & 0.5.10.post1 \\
\bottomrule
\end{tabular}
\caption{\label{tab:computing_infrastructure}
Computing infrastructure and software environment used for experiments.}

\end{table}

\section{Reproducibility Details}
All experiments were conducted using a combination of API-based inference and local model inference. Closed-source models were accessed through their official APIs. Open-source models were deployed locally using GPU-based inference. Table~\ref{tab:computing_infrastructure} summarizes the hardware and software configurations used for local inference and evaluation. For local experiments involving stochastic components, random seeds were fixed to 42 to ensure reproducibility.

\section{Alternative Pathway-Level Metrics}
\label{sec:appendix_metrics}
\subsection{OT-based Measurement}
We model pathway alignment as a global matching problem using OT, which establishes correspondences between steps by considering their overall matching cost without enforcing strict sequential alignment. Given a step-level consistency matrix $C\in \mathbb{R}^{m\times n}$, we convert it into a consistency-based distance matrix $D^{con}\in \mathbb{R}^{m\times n}$ by defining $D^{con}_{i,j}=1-C_{i,j}$. Since the OT formulation does not explicitly model the ordering of steps, we introduce a relative positional penalty $D^{pos}\in \mathbb{R}^{m\times n}$ to encourage order-aware alignment. This penalty is defined as the absolute difference between the normalized positions of the $i$-th reference step and the $j$-th predicted step:
\begin{equation}
    D^{pos}_{i,j}=\left| \frac{i-0.5}{m}-\frac{j-0.5}{n} \right|.
\end{equation}
We then combine the consistency-based distance and positional penalty into a unified transport cost matrix:
\begin{equation}
    D_{i,j}=\frac{\alpha D^{con}_{i,j}+\beta D^{pos}_{i,j}}{\alpha+\beta},
\end{equation}
where $\alpha=1.0$ controls the contribution of the consistency-based distance, while $\beta$ controls the contribution of positional alignment and is set to $0.2$ for GPT-5.4 Mini and Qwen3-8B, and $0.1$ for BERTScore.

We define a transport plan $T\in \mathbb{R}^{m\times n}$, where each entry $T_{i,j}$ represents the amount of mass transported from the $i$-th reference step to the $j$-th predicted step. Let $\mathbf{a}\in\mathbb{R}^m$ and $\mathbf{b}\in \mathbb{R}^{n}$ denote the marginal distributions over reference and predicted steps, respectively, which we set to uniform distributions, i.e., $\mathbf{a}_i=\frac{1}{m}$ and $\mathbf{b}_j=\frac{1}{n}$. Instead of standard balanced OT, we adopt an unbalanced OT formulation, which relaxes the exact marginal constraints and allows partial mass variation during transport, making it more suitable for pathway alignment. The resulting optimization objective is to find a transport plan that minimizes the total transport cost while softly enforcing the marginal constraints. Specifically, we adopt a KL-regularized formulation \cite{UOT1, UOT2}:
\begin{equation}
\begin{aligned}
\min_{T \ge 0} &\langle T, D \rangle_F+\lambda H(T)\\&+\tau_a \mathrm{KL}(T\mathbf{1}_n,\mathbf{a})+\tau_b \mathrm{KL}(T^\top \mathbf{1}_m,\mathbf{b}),
\end{aligned}
\end{equation}
where $H(T)$ is an entropic regularization term, the KL divergence terms softly penalize deviations from the marginal distributions $\mathbf{a}$ and $\mathbf{b}$, and $\lambda$, $\tau_a$ and $\tau_b$ control the strength of entropy regularization and marginal relaxation.

This formulation captures global alignment between pathways by allowing flexible many-to-many correspondences while softly encouraging positional consistency. As the final metric, we compute the pathway consistency cost induced by the optimal transport plan as
\begin{equation}
    \mathrm{PCC}_{\mathrm{OT}} = \frac{\langle T, D \rangle_F}{\|T\|_1}.
\end{equation}
Here, $\langle T, D \rangle_F$ denotes the total transport cost weighted by the transport plan, and $\|T\|_1$ normalizes the cost by the total transported mass. Lower $\mathrm{PCC}_{\mathrm{OT}}$ indicates better alignment between the reference CDP and the predicted CDP.

\subsection{LNDS-based Measurement}
In addition to global alignment strategies, we consider a lightweight approach that directly evaluates order consistency using the Longest Non-Decreasing Subsequence (LNDS). Unlike DTW and OT, which allow flexible matching between steps, this approach focuses on identifying the largest subset of matched steps that can be aligned while strictly preserving the decision order.

Let $s_i^*$ and $s_j$ denote the $i$-th reference step and the $j$-th predicted step from the reference CDP and predicted CDP, respectively. Starting from the step-level consistency matrix $C\in \mathbb{R}^{m\times n}$, we derive a candidate alignment for each reference step $s^*_i$ by selecting the predicted step index $j_i$ with the highest consistency score. As described in Section~\ref{sec:threshold_analysis}, we introduce a threshold $\gamma$ to avoid spurious matches caused by low-confidence alignments, and retain only matches satisfying $C_{i,j_i}\ge \gamma$. Reference steps below the threshold are treated as unmatched. This produces a sequence of matched predicted-step indices ordered by the reference CDP. We then compute the longest non-decreasing subsequence of this sequence, which represents the largest subset of matched steps that preserves the expected decision order.

The resulting subsequence length reflects how consistently the predicted CDP follows the correct decision order. As the final metric, we normalize the subsequence length as
\begin{equation}
    \mathrm{PCS}_{\mathrm{LNDS}} = \frac{m'}{m},
\end{equation}
where $m'$ denotes the length of the longest non-decreasing subsequence and $m$ denotes the number of reference steps. A higher $\mathrm{PCS}_{\mathrm{LNDS}}$ indicates better adherence to the guideline-defined decision order.

\begin{figure}[t]
    \centering
    \includegraphics[width=\linewidth]{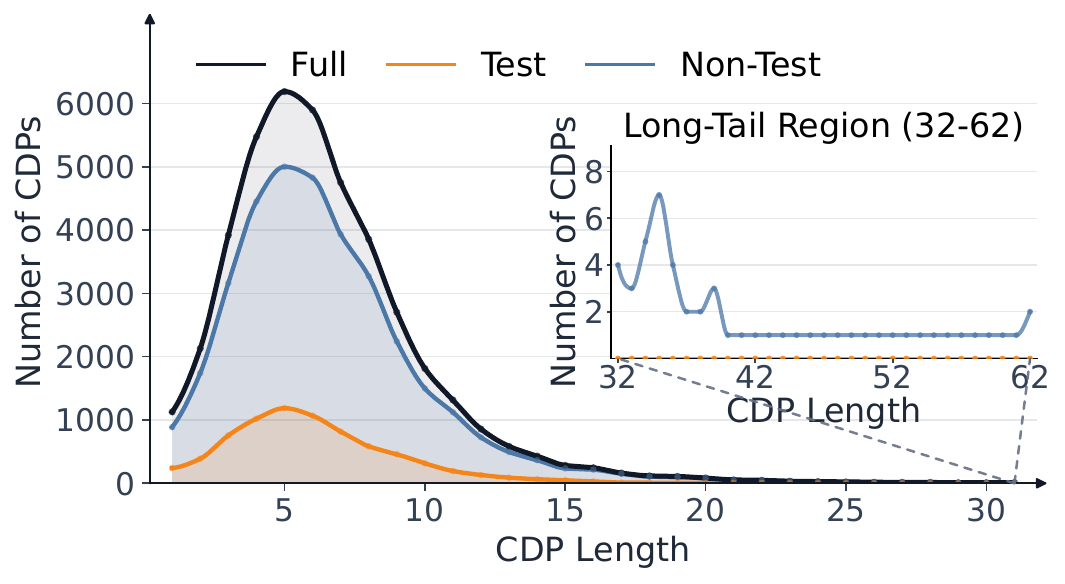}
    \caption{Distribution of reference CDP lengths. The inset shows a magnified view of the long-tail region.}
    \label{fig:cdp_length_distribution}
\end{figure}

\begin{table}[t]
\centering
\small
\begin{tabular}{l|c|c|c}
\toprule
Statistic & Full & Test & Non-Test \\
\midrule
Mean & 6.51 & 6.12 & 6.59 \\
Std. Dev. & 3.74 & 3.18 & 3.84 \\
Min & 1& 1 & 1 \\
P5 & 2& 2 & 2 \\
P25 & 4& 4 & 4 \\
Median & 6& 6 & 6 \\
P75 & 8& 8 & 8 \\
P95 & 13& 12 & 13 \\
Max & 62& 31 & 62 \\
\bottomrule
\end{tabular}  
\caption{\label{tab:cdp_length_statistics}
Summary statistics of guideline-defined CDP lengths across dataset splits.}
\end{table}

\section{Statistics of CDP Lengths}
We analyze the statistics and distribution of reference CDP lengths to characterize the structural properties of the guideline-defined CDPs. Here, CDP length is defined as the number of non-leaf nodes along a root-to-leaf pathway, corresponding to the number of clinical decision conditions excluding the final decision outcome. Figure~\ref{fig:cdp_length_distribution} presents the distribution of reference CDP lengths across the full CDP set and its test and non-test subsets, with the long-tail region magnified in the inset. Table~\ref{tab:cdp_length_statistics} summarizes the descriptive statistics of reference CDP lengths, including the mean, standard deviation, percentile statistics, and range. The test and non-test subsets exhibit similar CDP length statistics, suggesting that the held-out test set maintains similar structural characteristics to the overall CDP set.

\begin{table*}[t]
\centering
\small
\begin{tabular}{l|ccc}
\toprule
\multirow{2}{*}{Model} & \multicolumn{3}{c}{DTW-PCC (\%) $\downarrow$} \\
& GPT-5.4 Mini & Qwen3-8B & BERTScore \\
\midrule

\rowcolor{gray!15}
\multicolumn{4}{c}{\textit{Single-Turn Vignette Setting}} \\
Gemini-3.1-Pro-Preview&44.61 (44.07--45.16)&33.67 (33.16--34.19)&71.37 (71.13--71.62) \\
GPT-5.4&28.41 (27.94--28.86)&18.49 (18.08--18.88)&62.88 (62.61--63.14) \\
Claude-Sonnet-4.6&32.42 (31.90--32.93)&24.17 (23.72--24.64)&68.35 (68.09--68.61) \\
Qwen3-8B&47.88 (47.41--48.35)&33.66 (33.24--34.09)&74.07 (73.84--74.29) \\
Qwen3-32B&44.31 (43.81--44.81)&30.59 (30.15--31.03)&77.35 (77.13--77.57) \\
Qwen3-235B-A22B&41.23 (40.73--41.74)&29.14 (28.71--29.58)&77.51 (77.27--77.75) \\
Qwen3-30B-A3B-Instruct-2507&34.29 (33.82--34.77)&22.99 (22.60--23.40)&70.91 (70.68--71.14) \\
Qwen3-30B-A3B-Thinking-2507&50.17 (49.63--50.72)&37.37 (36.84--37.89)&73.23 (72.99--73.47) \\
Qwen3-Next-80B-A3B-Instruct&37.26 (36.78--37.74)&25.20 (24.79--25.60)&71.49 (71.24--71.74) \\
Qwen3-Next-80B-A3B-Thinking&51.31 (50.76--51.85)&39.26 (38.72--39.81)&74.65 (74.40--74.90) \\
DeepSeek-R1-Distill-Qwen-32B&36.98 (36.49--37.45)&25.95 (25.53--26.36)&72.03 (71.78--72.27) \\
gpt-oss-120b&39.24 (38.74--39.74)&26.63 (26.17--27.09)&74.45 (74.19--74.71) \\
LLaMA-3.3-70B-Instruct&34.71 (34.24--35.17)&25.24 (24.82--25.66)&70.79 (70.60--70.98) \\
HuatuoGPT-o1-7B&46.56 (46.09--47.01)&32.82 (32.40--33.24)&78.97 (78.79--79.16) \\
HuatuoGPT-o1-72B&33.87 (33.43--34.30)&23.32 (22.94--23.70)&71.09 (70.86--71.32) \\
Baichuan-M3-235B&42.35 (41.83--42.88)&29.80 (29.32--30.28)&73.91 (73.66--74.17) \\

\midrule
\rowcolor{gray!15}
\multicolumn{4}{c}{\textit{Multi-Turn Interactive Setting}} \\
Gemini-3.1-Pro-Preview&65.32 (64.67--65.96)&58.93 (58.22--59.66)&75.47 (75.03--75.90) \\
GPT-5.4&67.43 (66.79--68.06)&61.42 (60.70--62.14)&77.61 (77.19--78.03) \\
Claude-Sonnet-4.6&55.92 (55.38--56.46)&46.15 (45.58--46.74)&70.52 (70.18--70.86) \\
Qwen3-8B&64.34 (63.85--64.82)&53.63 (53.07--54.17)&73.84 (73.54--74.14) \\
Qwen3-32B&60.45 (59.95--60.95)&47.91 (47.37--48.43)&74.22 (73.92--74.53) \\
Qwen3-235B-A22B&63.66 (63.12--64.19)&54.69 (54.10--55.28)&74.64 (74.32--74.97) \\
Qwen3-30B-A3B-Instruct-2507&61.95 (61.44--62.45)&48.81 (48.28--49.35)&74.15 (73.85--74.45) \\
Qwen3-30B-A3B-Thinking-2507&72.61 (72.07--73.14)&65.50 (64.85--66.14)&80.02 (79.65--80.40) \\
Qwen3-Next-80B-A3B-Instruct&58.27 (57.77--58.76)&47.30 (46.77--47.85)&75.88 (75.58--76.18) \\
Qwen3-Next-80B-A3B-Thinking&69.15 (68.58--69.73)&59.45 (58.77--60.11)&79.96 (79.61--80.32) \\
DeepSeek-R1-Distill-Qwen-32B&82.12 (81.62--82.62)&78.80 (78.23--79.38)&88.31 (88.01--88.61) \\
gpt-oss-120b&64.78 (64.25--65.30)&56.81 (56.22--57.40)&77.50 (77.19--77.82) \\
LLaMA-3.3-70B-Instruct&61.04 (60.57--61.52)&46.84 (46.34--47.32)&82.32 (82.10--82.53) \\
HuatuoGPT-o1-7B&91.73 (91.32--92.15)&91.44 (91.00--91.88)&96.78 (96.60--96.96) \\
HuatuoGPT-o1-72B&70.28 (69.70--70.85)&63.12 (62.44--63.79)&83.27 (82.92--83.62) \\
Baichuan-M3-235B&68.27 (67.84--68.70)&54.12 (53.66--54.60)&81.16 (80.96--81.36) \\

\bottomrule
\end{tabular}
\caption{\label{tab:appendix_dtw_pcc_bootstrap}
DTW-based PCC results with bootstrap confidence intervals. Results are reported across different judge models under both single-turn vignette and multi-turn interactive settings. Each entry shows the mean PCC value with its 95\% bootstrap confidence interval. Lower values indicate better pathway consistency.}

\end{table*}

\begin{table*}[t]
\centering
\small
\begin{tabular}{l|ccc}
\toprule
\multirow{2}{*}{Model} & \multicolumn{3}{c}{OT-PCC (\%) $\downarrow$} \\
& GPT-5.4 Mini & Qwen3-8B & BERTScore \\
\midrule

\rowcolor{gray!15}
\multicolumn{4}{c}{\textit{Single-Turn Vignette Setting}} \\
Gemini-3.1-Pro-Preview&39.37 (38.88--39.86)&31.41 (30.96--31.87)&73.10 (72.86--73.35) \\
GPT-5.4&22.02 (21.62--22.40)&15.35 (15.02--15.66)&64.44 (64.21--64.69) \\
Claude-Sonnet-4.6&27.11 (26.68--27.55)&21.28 (20.90--21.67)&69.70 (69.45--69.95) \\
Qwen3-8B&36.94 (36.56--37.33)&25.72 (25.38--26.06)&74.22 (74.00--74.44) \\
Qwen3-32B&34.51 (34.10--34.91)&24.07 (23.72--24.42)&78.27 (78.06--78.47) \\
Qwen3-235B-A22B&33.94 (33.55--34.35)&24.92 (24.56--25.28)&78.89 (78.65--79.12) \\
Qwen3-30B-A3B-Instruct-2507&28.73 (28.35--29.12)&19.99 (19.67--20.33)&71.49 (71.28--71.70) \\
Qwen3-30B-A3B-Thinking-2507&41.57 (41.09--42.07)&31.10 (30.64--31.53)&73.98 (73.75--74.21) \\
Qwen3-Next-80B-A3B-Instruct&29.99 (29.60--30.37)&20.63 (20.32--20.95)&71.57 (71.33--71.79) \\
Qwen3-Next-80B-A3B-Thinking&42.75 (42.25--43.25)&32.82 (32.35--33.30)&75.83 (75.58--76.09) \\
DeepSeek-R1-Distill-Qwen-32B&30.35 (29.96--30.75)&21.52 (21.19--21.87)&73.78 (73.54--74.02) \\
gpt-oss-120b&31.18 (30.76--31.61)&21.58 (21.21--21.96)&76.04 (75.78--76.29) \\
LLaMA-3.3-70B-Instruct&29.23 (28.84--29.63)&22.23 (21.86--22.59)&72.60 (72.42--72.78) \\
HuatuoGPT-o1-7B&39.64 (39.24--40.04)&28.24 (27.88--28.61)&80.86 (80.68--81.04) \\
HuatuoGPT-o1-72B&27.41 (27.07--27.76)&19.44 (19.15--19.73)&72.24 (72.03--72.43) \\
Baichuan-M3-235B&34.52 (34.08--34.98)&24.73 (24.34--25.13)&74.84 (74.59--75.08) \\

\midrule
\rowcolor{gray!15}
\multicolumn{4}{c}{\textit{Multi-Turn Interactive Setting}} \\
Gemini-3.1-Pro-Preview&58.12 (57.39--58.85)&53.28 (52.51--54.05)&75.15 (74.69--75.60) \\
GPT-5.4&60.41 (59.69--61.12)&56.13 (55.35--56.89)&77.14 (76.69--77.57) \\
Claude-Sonnet-4.6&46.63 (46.06--47.19)&38.95 (38.36--39.55)&70.03 (69.68--70.37) \\
Qwen3-8B&55.19 (54.67--55.73)&46.13 (45.54--46.72)&73.50 (73.18--73.83) \\
Qwen3-32B&50.91 (50.38--51.45)&40.08 (39.52--40.63)&73.98 (73.65--74.30) \\
Qwen3-235B-A22B&55.69 (55.11--56.26)&47.90 (47.27--48.51)&74.43 (74.08--74.78) \\
Qwen3-30B-A3B-Instruct-2507&52.40 (51.83--52.96)&40.81 (40.24--41.36)&74.56 (74.24--74.88) \\
Qwen3-30B-A3B-Thinking-2507&65.82 (65.20--66.43)&60.00 (59.27--60.72)&79.80 (79.40--80.20) \\
Qwen3-Next-80B-A3B-Instruct&48.47 (47.96--48.99)&39.46 (38.90--40.02)&75.90 (75.57--76.21) \\
Qwen3-Next-80B-A3B-Thinking&61.78 (61.13--62.44)&53.25 (52.51--53.97)&79.93 (79.56--80.31) \\
DeepSeek-R1-Distill-Qwen-32B&78.70 (78.11--79.27)&75.61 (74.96--76.27)&89.41 (89.11--89.70) \\
gpt-oss-120b&56.01 (55.44--56.58)&49.52 (48.88--50.16)&77.62 (77.29--77.95) \\
LLaMA-3.3-70B-Instruct&50.51 (50.00--51.02)&37.85 (37.33--38.35)&84.01 (83.78--84.24) \\
HuatuoGPT-o1-7B&90.59 (90.12--91.06)&90.55 (90.07--91.04)&97.34 (97.17--97.50) \\
HuatuoGPT-o1-72B&63.91 (63.26--64.56)&57.93 (57.18--58.67)&84.06 (83.71--84.43) \\
Baichuan-M3-235B&56.35 (55.87--56.83)&43.05 (42.58--43.55)&82.62 (82.40--82.84) \\

\bottomrule
\end{tabular}
\caption{\label{tab:appendix_ot_pcc_bootstrap}
OT-based PCC results with bootstrap confidence intervals. Results are reported across different judge models under both single-turn vignette and multi-turn interactive settings. Each entry shows the mean PCC value with its 95\% bootstrap confidence interval. Lower values indicate better pathway consistency.}

\end{table*}

\begin{table*}[t]
\centering
\small
\begin{tabular}{l|ccc}
\toprule
\multirow{2}{*}{Model} & \multicolumn{3}{c}{LNDS-PCS (\%) $\uparrow$} \\
& GPT-5.4 Mini & Qwen3-8B & BERTScore \\
\midrule

\rowcolor{gray!15}
\multicolumn{4}{c}{\textit{Single-Turn Vignette Setting}} \\
Gemini-3.1-Pro-Preview&53.01 (52.37--53.65)&50.97 (50.35--51.58)&67.09 (66.48--67.68) \\
GPT-5.4&74.09 (73.55--74.65)&75.86 (75.32--76.40)&81.73 (81.28--82.17) \\
Claude-Sonnet-4.6&67.92 (67.32--68.52)&64.87 (64.30--65.45)&74.17 (73.64--74.71) \\
Qwen3-8B&56.20 (55.65--56.76)&57.58 (57.05--58.11)&67.22 (66.66--67.77) \\
Qwen3-32B&61.72 (61.14--62.28)&62.64 (62.09--63.17)&65.52 (64.92--66.11) \\
Qwen3-235B-A22B&59.99 (59.39--60.59)&58.56 (58.00--59.11)&59.89 (59.22--60.56) \\
Qwen3-30B-A3B-Instruct-2507&68.03 (67.46--68.58)&67.62 (67.08--68.17)&74.30 (73.80--74.80) \\
Qwen3-30B-A3B-Thinking-2507&47.48 (46.85--48.08)&49.60 (49.04--50.18)&63.42 (62.83--64.00) \\
Qwen3-Next-80B-A3B-Instruct&65.28 (64.71--65.85)&65.62 (65.07--66.15)&71.97 (71.43--72.51) \\
Qwen3-Next-80B-A3B-Thinking&45.97 (45.35--46.58)&47.53 (46.93--48.13)&58.35 (57.72--59.00) \\
DeepSeek-R1-Distill-Qwen-32B&62.89 (62.32--63.50)&60.98 (60.44--61.55)&69.41 (68.84--70.01) \\
gpt-oss-120b&59.69 (59.10--60.27)&63.09 (62.52--63.66)&64.39 (63.77--65.02) \\
LLaMA-3.3-70B-Instruct&67.53 (66.97--68.11)&62.50 (61.92--63.09)&76.22 (75.75--76.70) \\
HuatuoGPT-o1-7B&60.69 (60.11--61.27)&56.73 (56.17--57.31)&65.33 (64.72--65.93) \\
HuatuoGPT-o1-72B&72.97 (72.45--73.48)&71.75 (71.24--72.25)&78.62 (78.15--79.07) \\
Baichuan-M3-235B&58.18 (57.55--58.78)&59.19 (58.61--59.78)&65.11 (64.49--65.72) \\

\midrule
\rowcolor{gray!15}
\multicolumn{4}{c}{\textit{Multi-Turn Interactive Setting}} \\
Gemini-3.1-Pro-Preview&35.96 (35.24--36.68)&34.54 (33.80--35.29)&49.06 (48.25--49.89) \\
GPT-5.4&31.85 (31.16--32.55)&30.83 (30.12--31.56)&45.27 (44.46--46.09) \\
Claude-Sonnet-4.6&45.15 (44.51--45.81)&44.57 (43.88--45.23)&61.88 (61.25--62.52) \\
Qwen3-8B&36.90 (36.30--37.48)&37.26 (36.63--37.89)&57.69 (57.04--58.33) \\
Qwen3-32B&43.82 (43.19--44.44)&45.12 (44.47--45.76)&59.43 (58.79--60.04) \\
Qwen3-235B-A22B&36.74 (36.11--37.38)&35.32 (34.68--35.97)&54.73 (54.03--55.41) \\
Qwen3-30B-A3B-Instruct-2507&44.57 (43.95--45.21)&48.17 (47.53--48.81)&60.07 (59.48--60.68) \\
Qwen3-30B-A3B-Thinking-2507&25.90 (25.32--26.49)&25.84 (25.23--26.47)&41.66 (40.91--42.43) \\
Qwen3-Next-80B-A3B-Instruct&47.60 (46.98--48.24)&47.11 (46.42--47.79)&57.49 (56.83--58.14) \\
Qwen3-Next-80B-A3B-Thinking&32.98 (32.31--33.66)&35.16 (34.46--35.87)&45.61 (44.86--46.38) \\
DeepSeek-R1-Distill-Qwen-32B&16.45 (15.92--17.01)&14.24 (13.72--14.74)&27.51 (26.74--28.29) \\
gpt-oss-120b&34.23 (33.63--34.84)&32.89 (32.29--33.52)&49.52 (48.82--50.22) \\
LLaMA-3.3-70B-Instruct&47.70 (47.09--48.31)&50.42 (49.80--51.04)&49.64 (49.01--50.28) \\
HuatuoGPT-o1-7B&7.73 (7.29--8.17)&5.37 (5.01--5.74)&6.84 (6.39--7.29) \\
HuatuoGPT-o1-72B&33.17 (32.46--33.88)&30.94 (30.23--31.66)&38.82 (38.04--39.60) \\
Baichuan-M3-235B&52.89 (52.29--53.50)&57.37 (56.75--57.98)&58.18 (57.60--58.77) \\

\bottomrule
\end{tabular}
\caption{\label{tab:appendix_lnds_pcs_bootstrap}
LNDS-based PCS results with bootstrap confidence intervals. Results are reported across different judge models under both single-turn vignette and multi-turn interactive settings. Each entry shows the mean PCS value with its 95\% bootstrap confidence interval. Higher values indicate better pathway consistency.}

\end{table*}

\section{Additional Evaluation Results}
Tables~\ref{tab:appendix_dtw_pcc_bootstrap}, \ref{tab:appendix_ot_pcc_bootstrap}, and \ref{tab:appendix_lnds_pcs_bootstrap} report additional pathway consistency results under different pathway-level metrics and judge models, with 95\% confidence intervals estimated using 10,000 bootstrap resampling iterations. Across different judge models, similar model rankings and relative performance trends are generally preserved, suggesting that the pathway consistency evaluation is robust to variations in judge models. Similarly, DTW-PCC, OT-PCC, and LNDS-PCS exhibit consistent relative trends across models under both task settings, indicating that different pathway-level metrics provide broadly consistent assessments despite their different mathematical formulations. Moreover, the narrow confidence intervals further indicate that the estimated pathway consistency scores are stable.



\section{Prompt Templates}
\label{sec:appendix_prompts}
This section provides the English prompt templates used in MEGA-CDP. For readability, we use placeholders such as \texttt{\{guideline\}} and \texttt{\{case\_vignette\}} to indicate task-specific inputs. Tables~\ref{tab:prompt1}--\ref{tab:prompt9} present the complete prompt templates in their original wording.

\clearpage
\onecolumn
{\small
\begin{longtable}{p{0.96\textwidth}}
\toprule
\textbf{Prompt 1: Reasoning Tree Extraction} \\
\midrule
\endfirsthead

\toprule
\textbf{Prompt 1: Reasoning Tree Extraction (continued)} \\
\midrule
\endhead

You are a medical expert. You will receive a clinical guideline as input. Your task is to read the input, understand each paragraph, extract all content related to clinical decision-making, and convert it into a decision tree represented as a JSON object. Each non-leaf node must branch into two paths, "yes" and "no", and represent a criteria-check question about patient-specific clinical data or findings (e.g., “Is symptom A present?”, “Is lab value B $\ge$ threshold?”, “Did event C occur within time window D?”); each leaf node should represent a specific conclusion. \\
\\
Before constructing the decision tree, analyze the entire guideline and determine whether the primary clinical focus for the selected disease is diagnosis, classification, screening, or management. Report this explicitly in the final JSON under "primary\_clinical\_focus". Use this determination to guide node construction, but do not represent it as a decision node. The decision tree must represent a complete and clinically meaningful pathway. It should reflect the main logical architecture of the selected domain, typically with multiple hierarchical levels, unless the guideline itself is inherently simple.\\
\\
When constructing the decision tree:\\
- Replace vague or implicit conditions with explicit, objective, and clinically actionable criteria (e.g., symptoms, laboratory thresholds, imaging findings, or time-based conditions), strictly based on the guideline text.\\
- Each decision node should assess only one independent clinical variable or threshold. Avoid merging multiple distinct criteria into a single question. When multiple criteria are required by the guideline, represent them as sequential hierarchical nodes rather than compound logical expressions.\\
- Do NOT introduce external medical knowledge or infer criteria not explicitly stated in the guideline. If the guideline text is ambiguous, preserve the ambiguity without speculation.\\
- Do NOT include nodes that pose questions about clinical goals, tasks, intentions, scopes, or diagnostic objectives (e.g., whether the clinical task is to diagnose, classify, screen, or manage a condition). All decision nodes must be based solely on patient-specific clinical data or findings.\\
- Do NOT create decision nodes that ask whether a diagnosis/classification/screening result/management decision is already established or “clear/confirmed/likely” (e.g., “Is the diagnosis clear?”, “Has the patient been diagnosed with…?”, “Is the patient classified as…?”, “Should treatment be started?”). Decision nodes must test only patient-level observable criteria (symptoms, signs, labs, imaging, events) rather than the existence of a clinical conclusion.\\
- Do NOT create redundant or logically equivalent nodes. If a condition has already been fully determined by a previous yes/no branch (e.g., the "no" branch of "Is the patient pregnant?" already implies the patient is not pregnant), do not restate the complementary condition as a new decision node.\\
- If the guideline uses conclusion-like phrasing (e.g., “diagnose X if…”, “classify as Y when…”, “screen positive if…”, “treat if…”), convert it into explicit, verifiable criteria nodes (the “if/when” conditions), and place the conclusion itself only in a leaf node.\\
\\
In addition to the decision tree, identify the guideline evidence supporting each node:\\
- Assign a unique numeric index to every node in the decision tree.\\
- For each node, identify and quote the exact sentence(s) from the guideline that justify the node. Quoted evidence must consist of complete declarative sentences containing clinical statements or recommendations from the guideline body, and must not be drawn from section titles, headings, subheadings, interrogative statements, table/figure captions or other non-informative content.\\
- All quoted sentences must be reproduced in full, without omission or truncation.\\
- All quoted guideline sentences must pertain specifically to the selected disease and population. Do not cite general background statements or recommendations that are not explicitly tied to the selected focus.\\
- Allow one-to-many or many-to-one mappings between nodes and guideline sentences.\\
\\
Note: The guideline text is automatically converted from a PDF into Markdown format. As a result, it may contain structural inconsistencies, incorrect heading levels, formatting artifacts, duplicated text, table fragments, or other non-informative content. Do NOT rely solely on heading hierarchy or formatting to determine logical structure. Instead, identify and extract clinically meaningful, decision-relevant statements based on their semantic content. Ignore irrelevant, repetitive, or non-clinical material that does not contribute to patient-specific clinical decision-making. Do not assume that structurally well-formatted or clearly segmented subsections represent the primary clinical objective of the guideline. Selection of the focus must be based on semantic centrality and scope rather than formatting clarity or local structural neatness.\\
\\
Do not include any explanations, commentary, or content beyond the specified JSON output.\\
\\
\#\# Guideline:\\
\{Guideline\}\\
\\
\#\# Example of a Leaf-Node JSON Format:\\
\{\\
\hspace{2em}"index": 1,\\
\hspace{2em}"leaf": true,\\
\hspace{2em}"conclusion": "a specific conclusion"\\
\}\\
\\
\#\# Example of a Non-Leaf-Node JSON Format:\\
\{\\
\hspace{2em}"index": 1,\\
\hspace{2em}"description": "description of a specific decision step",\\
\hspace{2em}"yes": \{\},\\
\hspace{2em}"no": \{\}\\
\}
\\
\#\# Output Format:\\
\{\\
\hspace{2em}"primary\_clinical\_focus": "diagnosis | classification | screening | management",\\
\hspace{2em}"decision\_tree": \{\\
\hspace{4em}"index": 1,\\
\hspace{4em}...\\
\hspace{2em}\},\\
\hspace{2em}"node\_guideline\_mapping": \{\\
\hspace{4em}"1": [\\
\hspace{6em}"Quoted guideline sentence 1",\\
\hspace{6em}"Quoted guideline sentence 2"\\
\hspace{4em}],\\
\hspace{4em}"2": [\\
\hspace{6em}...\\
\hspace{4em}]\\
\hspace{2em}\}\\
\}\\
\\
\#\# Your Output:\\
\bottomrule
\caption{\label{tab:prompt1}
Prompt template for reasoning tree extraction.}
\\
\end{longtable}
}

\clearpage
{\small
\begin{longtable}{p{0.96\textwidth}}
\toprule
\textbf{Prompt 2: Case Vignette Synthesis} \\
\midrule
\endfirsthead

\toprule
\textbf{Prompt 2: Case Vignette Synthesis (continued)} \\
\midrule
\endhead
You are a medical expert. I will provide you with a clinical decision path consisting of multiple steps. Each step includes a "condition" representing the clinical condition or decision criterion, and a "decision\_outcome" indicating whether the condition is satisfied. In addition, I will specify the primary clinical focus of this path (e.g., diagnosis, classification, screening, or management). Your task is to convert this clinical decision path into a single, coherent, and natural-sounding clinical case description using medically styled language, appropriate to the specified primary clinical focus. The output should be written as a realistic clinical case vignette, resembling the narrative style commonly used in medical textbooks or case reports, rather than a summary or explanation of a decision pathway.\\
\\
Note the following constraints:\\
1. The tone of each step's rewritten content must reflect its decision\_outcome value. Use an affirmative sentence if the decision\_outcome is “yes”, and use a negative sentence if the decision\_outcome is “no”.\\
2. For any medical indicator involving a numerical range, you must provide a specific numeric value. The value must fall within the range if the decision\_outcome indicates the condition is satisfied, and outside the range if the decision\_outcome indicates the condition is not satisfied.\\
3. If the same medical indicator appears in multiple steps, ensure that the chosen numeric value is consistent and falls within the intersection of all applicable ranges.\\
4. All numeric values chosen for medical indicators must be clinically reasonable and plausible within real-world medical contexts.\\
5. If a step requires multiple conditions to be met simultaneously (logical AND), the description must fully and accurately reflect all such conditions.\\
6. If a step represents partially satisfied conditions (logical OR), you may select one or more logically valid conditions to include in the description, as long as they remain clinically consistent.\\
7. The description should integrate information from all steps naturally into a single narrative, without explicitly referencing decision steps, pathway logic, or the existence of a clinical pathway. Do not include step numbers or labels.\\
8. Write in the past tense as if describing a patient's clinical course.\\
9. Avoid gendered pronouns (e.g., “he”, “she”) when referring to the patient, unless the clinical condition explicitly involves sex-specific characteristics. If necessary, refer to the patient using neutral terms such as “the patient” or omit pronouns entirely.\\
10. The final sentence of the output must explicitly and neutrally state the primary clinical focus of this pathway (i.e., diagnosis, classification, screening, or management).\\
11. Do not use wording that implies definitive diagnosis, risk stratification, or treatment decisions beyond what is appropriate for the specified primary clinical focus. Do not include meta-commentary or self-referential statements (e.g., “this description reflects…”, “this pathway focuses on…”), except for the required final sentence explicitly stating the primary clinical focus.\\
12. Do not add any extra information beyond what is provided.\\
\\
\#\# Primary Clinical Focus:\\
\{Primary\_Clinical\_Focus\}\\
\\
\#\# Clinical Decision Path:\\
\{Condition\_Path\}\\
\\
\#\# Your Output:\\
\bottomrule
\caption{\label{tab:prompt2}
Prompt template for case vignette synthesis.}
\\
\end{longtable}
}

\clearpage
{\small
\begin{longtable}{p{0.96\textwidth}}
\toprule
\textbf{Prompt 3: Reasoning Narrative Conversion} \\
\midrule
\endfirsthead

\toprule
\textbf{Prompt 3: Reasoning Narrative Conversion (continued)} \\
\midrule
\endhead
You are a medical expert. You will be provided with a single JSON object containing information about a clinical decision pathway and a corresponding patient case under a specific clinical focus.\\
\\
The input JSON of the pathway contains:\\
- "primary\_clinical\_focus": the clinical purpose of the pathway,\\
- "decision\_path": an ordered list of decision steps along a single pathway,\\
- "case\_vignette": a narrative description of the patient’s clinical presentation.\\
\\
Each element in "decision\_path" represents one step along the pathway and may be one of the following:\\
- For non-final decision steps:\\
  - "condition": the clinical condition or decision question at this step,\\
  - "decision\_outcome": whether the condition is satisfied along this path ("yes" or "no"),\\
  - "guideline\_evidence": one or more supporting guideline statements.\\
\\
- For the final step:\\
  - "conclusion": the terminal clinical outcome of the pathway,\\
  - "guideline\_evidence": one or more supporting guideline statements.\\
\\
Your task is to produce a single JSON array of strings that represents the clinical decision reasoning along this pathway. The output list order must match the order of decision steps in the input "decision\_path".\\
\\
For each decision step except the final one:\\
- Integrate the relevant guideline evidence with the clinical context provided by the case vignette.\\
- When multiple guideline statements are present, identify the relevant decision logic.\\
- Synthesize this information into one concise and formal reasoning statement.\\
\\
The final step represents the terminal outcome of the pathway. Instead of step-level reasoning, provide a brief, single-sentence integrative statement that synthesizes the preceding decision reasoning and leads to the stated conclusion, without introducing new evidence or decision logic.\\
\\
Output must consist only of the JSON array, with no additional text.\\
\\
\#\# Input:\\
\{Input\}\\
\\
\#\# Output Format:\par
[\\
  "Reasoning statement for step 1 ...",\\
  "Reasoning statement for step 2 ...",\\
  "Reasoning statement for step 3 ..."\par
]\\
\\
\#\# Your Output:\\
\bottomrule
\caption{\label{tab:prompt3}
Prompt template for reasoning narrative conversion.}
\\
\end{longtable}
}

\clearpage
{\small
\begin{longtable}{p{0.96\textwidth}}
\toprule
\textbf{Prompt 4: Single-Turn Inference} \\
\midrule
\endfirsthead

\toprule
\textbf{Prompt 4: Single-Turn Inference (continued)} \\
\midrule
\endhead
You are a medical expert. I will provide you with a clinical guideline, a case vignette, and an example of the required output format. You must strictly follow the content of the guideline to make clinical decisions for the case, according to the specified primary clinical focus. Your output must be concise and formal, and must include both a structured clinical decision process and a final clinical outcome.\\
\\
In the clinical decision process:\\
- The process must be divided into clear, sequential steps.\\
- Each decision step must explicitly state the core reasoning logic that links the key patient symptoms or findings to that clinical decision step.\\
- Each decision step must assess only one independent clinical variable or threshold. Do not merge multiple distinct criteria into a single reasoning step.\\
\\
In the clinical decision outcome:\\
- If the guideline contains an explicit rule or conclusion that can be directly applied to the case for the specified primary clinical focus, you must state exactly one explicit clinical conclusion that appears in the guideline.\\
- Otherwise, you must state exactly that the guideline is not applicable.\\
- No alternative conclusions, probabilities, or additional explanations are allowed.\\
\\
Note: The guideline text was automatically converted from a PDF into Markdown format and may contain structural inconsistencies, incorrect heading levels, formatting artifacts, duplicated text, table fragments, or other non-informative content. Do NOT rely on heading hierarchy, formatting, or layout cues to determine logical structure. Instead, base all reasoning strictly on explicit, clinically meaningful statements within the guideline text. Ignore irrelevant, repetitive, non-clinical, or clearly corrupted formatting content. All reasoning must rely exclusively on explicitly stated content in the provided guideline text.\\
\\
The output must not contain any explanations, commentary, or content beyond what is shown in the provided example format, and it must strictly adhere to the format.\\
\\
\#\# Guideline:\\
\{Guideline\}\\
\\
\#\# Case Vignette:\\
\{Case\_Vignette\}\\
\\
\#\# Output Format:\\
\#\#\# Clinical Decision Process: \\
1. A core reasoning logic that links the key patient symptoms or findings to explicitly stated rules or thresholds in the clinical guideline.\\
2. ...\\
...\\
\#\#\# Clinical Decision Outcome: ...\\
\\
\#\# Your Output:\\
\bottomrule
\caption{\label{tab:prompt4}
Prompt template for single-turn inference.}
\\
\end{longtable}
}

\clearpage
{\small
\begin{longtable}{p{0.96\textwidth}}
\toprule
\textbf{Prompt 5: Single-Turn Step-Level Consistency Scoring} \\
\midrule
\endfirsthead

\toprule
\textbf{Prompt 5: Single-Turn Step-Level Consistency Scoring (continued)} \\
\midrule
\endhead
You are a medical expert.\\
\\
You will receive two clinical decision pathways:\\
\\
- Reference Clinical Pathway: a reference sequence of clinical reasoning steps\\
- Candidate Clinical Pathway: a pathway to be evaluated against the reference\\
\\
Each pathway is provided as a list of strings, where each string represents one clinical decision step.\\
\\
In the Reference Clinical Pathway, the steps are labeled sequentially from R1 to R\{Reference\_Pathway\_Length\}, for a total of \{Reference\_Pathway\_Length\} steps.\\
In the Candidate Clinical Pathway, the steps are labeled sequentially from C1 to C\{Candidate\_Pathway\_Length\}, for a total of \{Candidate\_Pathway\_Length\} steps.\\
\\
Your task is to compare every step in the Reference Clinical Pathway with every step in the Candidate Clinical Pathway and evaluate their consistency.\\
\\
Evaluation Criteria:\\
Consistency (1–5) – the degree to which the Candidate Step represents the same clinical decision or logical judgment as the Reference Step.\\
\\
Scoring Guide:\\
First read and understand each clinical pathway as a whole. The meaning of each step should be interpreted in the context of its pathway.\\
For each pair (Ri, Cj):\\
Identify the key decision points in the Reference Step (such as symptoms, test results, thresholds, clinical conditions, or required actions) and understand the decision logic expressed in that step within the context of the pathway.\\
Then evaluate whether the Candidate Step, when interpreted within its own pathway context, preserves the same decision meaning by covering these key decision points in a way that supports the same decision logic.\\
Matching concepts alone is not sufficient for a high score. A high score requires both strong coverage of the Reference Step’s key decision points and the same decision logic.\\
\\
Score Definitions:\\
5: The Candidate Step covers all key decision points in the Reference Step and expresses the same decision logic correctly.\\
4: The Candidate Step covers most key decision points in the Reference Step and preserves essentially the same decision logic, missing only minor details.\\
3: The Candidate Step overlaps with the Reference Step on some key decision points, but misses important information or only partially preserves the decision logic.\\
2: The Candidate Step mentions related concepts or partially related information, but represents a different decision logic, reasoning stage, or type of judgment.\\
1: The Candidate Step is unrelated to the key decision points and decision logic in the Reference Step, or expresses a clearly incorrect or contradictory condition.\\
\\
Output Format:\\
For each Reference step Ri, output all corresponding Candidate comparisons in ONE line.\\
Within each line, list all pairs (Ri, Cj) using the format, separated by commas:\\
R<i>-C1: <score>, R<i>-C2: <score>, ..., R<i>-C\{Candidate\_Pathway\_Length\}: <score>\\
\\
For example:\\
R1-C1: 5, R1-C2: 2\\
R2-C1: 3, R2-C2: 5\\
\\
Output Requirements:\\
- Output all \{Reference\_Pathway\_Length\} × \{Candidate\_Pathway\_Length\} step pairs exactly once.\\
- Output exactly \{Reference\_Pathway\_Length\} lines (one line per Reference step).\\
- Each line must contain exactly \{Candidate\_Pathway\_Length\} pairs, covering C1 to C\{Candidate\_Pathway\_Length\} in order.\\
- Each pair must follow the format R<i>-C<j>: <score>.\\
- Use only the provided labels: R1 to R\{Reference\_Pathway\_Length\} and C1 to C\{Candidate\_Pathway\_Length\}. Do not use any label outside these ranges.\\
- Within each line, pairs must be separated by a comma.\\
- Output only the pairwise scores. Do not include explanations, comments, headings, or additional text.\\
- Before producing the final output, verify that all possible Reference–Candidate pairs are included and that no pair is missing or duplicated.\\
\\
\#\# Reference Clinical Pathway:\\
\{Reference\_Pathway\}\\
\\
\#\# Candidate Clinical Pathway:\\
\{Candidate\_Pathway\}\\
\\
\#\# Your Output:\\
\bottomrule
\caption{\label{tab:prompt5}
Prompt template for single-turn step-level consistency scoring.}
\\
\end{longtable}
}

{\small
\begin{longtable}{p{0.96\textwidth}}
\toprule
\textbf{Prompt 6: Multi-Turn Inference} \\
\midrule
\endfirsthead

\toprule
\textbf{Prompt 6: Multi-Turn Inference (continued)} \\
\midrule
\endhead
You are a doctor. Your task is to determine the correct clinical conclusion strictly according to the provided guideline.\\
\\
You will receive the patient's clinical inquiry goal. Based on this goal, you must follow the explicit decision-making process in the guideline and ask the patient targeted questions step by step until either:\\
- exactly one clinical conclusion is determined, or\\
- it is impossible to determine a single conclusion based on the guideline and the patient's answers.\\
\\
Follow these rules:\\
1. Base all reasoning only on explicit statements in the guideline. Do not use external medical knowledge.\\
2. Ask only ONE question at a time. Each question must assess exactly ONE independent clinical criterion explicitly stated in the guideline.\\
3. Do not repeat questions. \\
4. Follow the guideline strictly in order. Do not skip, reorder, or jump between decision steps. Each question must correspond to the next immediate unresolved criterion based on the dialogue history.\\
5. Determine the next question based only on the guideline and the criteria already assessed in the dialogue. Treat all previously assessed criteria as fixed and do not reconsider or reinterpret them.\\
6. The patient can only answer "Yes" or "No". All questions must be answerable with "Yes" or "No". Convert any numerical threshold into a Yes/No question. If a Yes/No question could cause multiple interpretations or confusion, revise it so that it assesses exactly one criterion and can be answered clearly with "Yes" or "No".\\
7. Every criterion must be resolved as either "Yes" or "No". Do not assume missing or unavailable information.\\
8. Continue asking questions as long as there exists any unassessed criterion that could still change the final conclusion.\\
9. If the guideline leads to exactly one clinical conclusion based on the patient's answers, output it immediately.\\
10. When the guideline criteria strictly determine a single clinical conclusion based on the patient's answers, output exactly: 'Final Conclusion: <short clinical conclusion>'.\\
11. When no remaining unasked criterion can resolve the decision to a single conclusion, meaning the guideline does not contain an explicit rule or conclusion that can be directly applied to the patient's clinical inquiry goal, output exactly: 'Final Conclusion: The guideline does not contain an explicit rule or conclusion that can be directly applied based on the available patient information.'\\
12. Only output:\\
- one Yes/No question, OR\\
- the Final Conclusion in the exact specified format.\\
\\
Note: The guideline text was automatically converted from a PDF into Markdown format and may contain structural inconsistencies, incorrect heading levels, formatting artifacts, duplicated text, table fragments, or other non-informative content. Do NOT rely on heading hierarchy, formatting, or layout cues to determine logical structure. Instead, base all reasoning strictly on explicit, clinically meaningful statements within the guideline text. Ignore irrelevant, repetitive, non-clinical, or clearly corrupted formatting content. All reasoning must rely exclusively on explicitly stated content in the provided guideline text.\\
\\
Do not provide explanations, reasoning steps, or probabilities.\\
\\
\#\# Patient's Clinical Inquiry Goal\\
\{Clinical\_Inquiry\_Goal\}\\
\\
\#\# Guideline:\\
\{Guideline\}\\
\bottomrule
\caption{\label{tab:prompt6}
Prompt template for multi-turn inference.}
\\
\end{longtable}
}

\clearpage
{\small
\begin{longtable}{p{0.96\textwidth}}
\toprule
\textbf{Prompt 7: Multi-Turn Patient Environment} \\
\midrule
\endfirsthead

\toprule
\textbf{Prompt 7: Multi-Turn Patient Environment (continued)} \\
\midrule
\endhead
You will role-play as a patient. You must answer the doctor's questions as if you are a real person with a coherent and consistent underlying health state.\\
\\
Follow these rules:\\
1. Keep your answer to exactly one word: "Yes" or "No".\\
2. Do not add explanations or any additional information.\\
3. Use the Case Vignette as the foundation of your health state. All answers must be consistent with the information explicitly stated in the vignette.\\
4. If the vignette does not explicitly provide enough information:\\
- Assume a realistic and coherent patient profile that is fully consistent with the vignette.\\
- Answer based on this assumed patient.\\
5. Maintain strict internal consistency:\\
- Your answers must not contradict the vignette.\\
- Your answers must not contradict any of your previous answers.\\
6. Do not reveal that you are using a case vignette or an assumed patient.\\
\\
\#\# Case Vignette:\\
\{Case\_Vignette\}\\
\bottomrule
\caption{\label{tab:prompt7}
Prompt template for multi-turn patient environment.}
\\
\end{longtable}
}

\clearpage
{\small
\begin{longtable}{p{0.96\textwidth}}
\toprule
\textbf{Prompt 8: Multi-Turn Step-Level Consistency Scoring} \\
\midrule
\endfirsthead

\toprule
\textbf{Prompt 8: Multi-Turn Step-Level Consistency Scoring (continued)} \\
\midrule
\endhead
You are a medical expert.\\
\\
You will receive two clinical questioning pathways:\\
\\
- Reference Clinical Questioning Pathway: a reference sequence of clinician questions used to gather patient information\\
- Candidate Clinical Questioning Pathway: a questioning pathway to be evaluated against the reference\\
\\
Each pathway is provided as a list of strings, where each string represents one clinical question.\\
\\
In the Reference Clinical Questioning Pathway, the questions are labeled sequentially from R1 to R\{Reference\_Pathway\_Length\}, for a total of \{Reference\_Pathway\_Length\} questions.\\
In the Candidate Clinical Questioning Pathway, the questions are labeled sequentially from C1 to C\{Candidate\_Pathway\_Length\}, for a total of \{Candidate\_Pathway\_Length\} questions.\\
\\
Your task is to compare every question in the Reference Clinical Questioning Pathway with every question in the Candidate Clinical Questioning Pathway and evaluate their consistency.\\
\\
Evaluation Criteria:\\
Consistency (1–5) – the degree to which the Candidate Question reflects the same clinical intent, information need, or reasoning purpose as the Reference Question.\\
\\
Scoring Guide:\\
First read and understand each questioning pathway as a whole. Each question should be interpreted in the context of its pathway, including what has already been asked and what information is being pursued.\\
For each pair (Ri, Cj):\\
Identify the key information targets in the Reference Question (such as symptoms, duration, severity, location, timing, risk factors, medical history, test results, or clarifications being requested), and understand the intent of the question within the pathway.\\
Then evaluate whether the Candidate Question, when interpreted within its own pathway context, seeks the same information and serves the same clinical reasoning purpose.\\
Consider whether the two questions contribute to the same clinical reasoning process or information-gathering goal within their respective pathways.\\
Superficial keyword overlap is not sufficient. A high score requires both alignment in the information being requested and consistency in the clinical reasoning purpose.\\
\\
Score Definitions:\\
5: The Candidate Question covers all key information targets in the Reference Question and expresses the same questioning intent and clinical reasoning purpose.\\
4: The Candidate Question covers most key information targets and preserves essentially the same intent and reasoning purpose, missing only minor details.\\
3: The Candidate Question overlaps with the Reference Question on some information targets, but misses important aspects or only partially aligns with the intent.\\
2: The Candidate Question mentions related concepts or partially related information, but represents a different questioning intent, reasoning stage, or purpose.\\
1: The Candidate Question is unrelated to the information targets and intent of the Reference Question, or reflects a clearly different or inappropriate line of questioning.\\
\\
Output Format:\\
For each Reference question Ri, output all corresponding Candidate comparisons in ONE line.\\
Within each line, list all pairs (Ri, Cj) using the format, separated by commas:\\
R<i>-C1: <score>, R<i>-C2: <score>, ..., R<i>-C\{Candidate\_Pathway\_Length\}: <score>\\
\\
For Example:\\
R1-C1: 5, R1-C2: 2\\
R2-C1: 3, R2-C2: 5\\
\\
Output Requirements:\\
- Output all \{Reference\_Pathway\_Length\} × \{Candidate\_Pathway\_Length\} question pairs exactly once.\\
- Output exactly \{Reference\_Pathway\_Length\} lines (one line per Reference question).\\
- Each line must contain exactly \{Candidate\_Pathway\_Length\} pairs, covering C1 to C\{Candidate\_Pathway\_Length\} in order.
- ach pair must follow the format R<i>-C<j>: <score>.\\
- Use only the provided labels: R1 to R\{Reference\_Pathway\_Length\} and C1 to C\{Candidate\_Pathway\_Length\}. Do not use any label outside these ranges.\\
- Within each line, pairs must be separated by a comma.\\
- Output only the pairwise scores. Do not include explanations, comments, headings, or additional text.\\
- Before producing the final output, verify that all possible Reference–Candidate pairs are included and that no pair is missing or duplicated.\\
\\
\#\# Reference Clinical Questioning Pathway:\\
\{Reference\_Pathway\}\\
\\
\#\# Candidate Clinical Questioning Pathway:\\
\{Candidate\_Pathway\}\\
\\
\#\# Your Output:\\
\bottomrule
\caption{\label{tab:prompt8}
Prompt template for multi-turn step-level consistency scoring.}
\\
\end{longtable}
}

{\small
\begin{longtable}{p{0.96\textwidth}}
\toprule
\textbf{Prompt 9: Outcome Accuracy Scoring} \\
\midrule
\endfirsthead

\toprule
\textbf{Prompt 9: Outcome Accuracy Scoring (continued)} \\
\midrule
\endhead
You are a medical expert.\\
\\
You will receive two clinical conclusions:\\
- Reference Clinical Conclusion: a reference statement representing a final clinical judgment\\
- Candidate Clinical Conclusion: a conclusion to be evaluated against the reference\\
\\
Each clinical conclusion is provided as a text that represents the final outcome of clinical reasoning. It may consist of one or more sentences and may include a diagnosis, clinical assessment, or recommended management decision.\\
\\
Your task is to evaluate the consistency between these two conclusions.\\
\\
Evaluation Criteria:\\
Consistency (1–5) – the degree to which the Candidate Clinical Conclusion expresses the same clinical meaning as the Reference Clinical Conclusion.\\
\\
Evaluation Instructions:\\
1. Carefully read both conclusions.\\
2. Identify the key clinical elements in the Reference Clinical Conclusion.\\
3. Evaluate whether the Candidate Clinical Conclusion conveys the same clinical meaning.\\
4. Interpret both conclusions in a clinically precise manner; do not rely on superficial wording similarity.\\
\\
Score Definitions:\\
5: The two conclusions express the same clinical meaning with equivalent medical accuracy.\\
4: The conclusions are largely consistent, with only minor differences in wording or detail.\\
3: The conclusions overlap partially but differ in important aspects.\\
2: The conclusions mention related concepts but represent different clinical meanings.\\
1: The conclusions are contradictory or clinically unrelated.\\
\\
Output Requirements:\\
- Output only a single integer (1–5)\\
- Do not include explanations, comments, or additional text\\
\\
\#\# Reference Clinical Conclusion:\\
\{Reference\_Conclusion\}\\
\\
\#\# Candidate Clinical Conclusion:\\
\{Candidate\_Conclusion\}\\
\\
\#\# Your Output:\\
\bottomrule
\caption{\label{tab:prompt9}
Prompt template for outcome accuracy scoring.}
\\
\end{longtable}
}

\twocolumn

\end{document}